%% file: main.tex
\documentclass{article}

\usepackage{microtype}
\usepackage{graphicx}
\usepackage{subfigure}
\usepackage{booktabs} 

\usepackage{color}
\usepackage{enumitem}
\usepackage{multirow}
\usepackage{bbm}
\usepackage{duckuments}

\usepackage{hyperref}
\usepackage{listings}

\usepackage[accepted]{icml2023}

\usepackage{amsmath}
\usepackage{amssymb}
\usepackage{mathtools}
\usepackage{amsthm}

\usepackage[capitalize,noabbrev]{cleveref}

\theoremstyle{plain}

\theoremstyle{definition}

\theoremstyle{remark}

\usepackage[textsize=tiny]{todonotes}

\input{defs.tex}
\newcommand{\stdv}[1]{\scalebox{.70}{~$\pm$~#1}}

\icmltitlerunning{Aligning Text-to-Image Models using Human Feedback}

\begin{document}

\twocolumn[
\icmltitle{Aligning Text-to-Image Models using Human Feedback}



\icmlsetsymbol{equal}{*}

\begin{icmlauthorlist}
\icmlauthor{Kimin Lee}{yyy}
\icmlauthor{Hao Liu}{comp}
\icmlauthor{Moonkyung Ryu}{yyy}
\icmlauthor{Olivia Watkins}{comp}
\icmlauthor{Yuqing Du}{comp} \\
\icmlauthor{Craig Boutilier}{yyy}
\icmlauthor{Pieter Abbeel}{comp}
\icmlauthor{Mohammad Ghavamzadeh}{yyy}
\icmlauthor{Shixiang Shane Gu}{yyy}
\end{icmlauthorlist}

\icmlaffiliation{yyy}{Google Research}
\icmlaffiliation{comp}{University of California, Berkeley}

\icmlcorrespondingauthor{Kimin Lee}{kiminl@google.com}

\icmlkeywords{Machine Learning, ICML}

\vskip 0.3in
]



\printAffiliationsAndNotice{}  

\begin{abstract}
Deep generative models have shown impressive results in text-to-image synthesis.
However, current text-to-image models often generate images that are inadequately aligned with text prompts. We propose a fine-tuning method for aligning such models using human feedback, comprising three stages.
First, we collect human feedback assessing model output alignment from a set of diverse text prompts.
We then use the human-labeled image-text dataset to train a reward function
that predicts human feedback. Lastly, the text-to-image model is fine-tuned by maximizing \emph{reward-weighted} likelihood to improve image-text alignment.
Our method generates objects with specified colors, counts and backgrounds more accurately than the pre-trained model.
We also analyze several design choices and find that careful investigations on such design choices are important in balancing the alignment-fidelity tradeoffs.
Our results demonstrate the potential for learning from human feedback to significantly improve text-to-image models.
\end{abstract}

\input{sections/intro}
\input{sections/relatedwork}

\input{sections/method}

\input{sections/experiments}

\input{sections/conclusion}

\section*{Acknowledgements}

We thank Peter Anderson, Douglas Eck, Junsu Kim, Changyeon Kim, Jongjin Park, Sjoerd van Steenkiste, Younggyo Seo, and Guy Tennenholtz for providing helpful comments and suggestions.
Finally, we would like to thank Sehee Yang for providing valuable feedback on user interface and constructing the initial version of human data, without which this project would not have been possible.

\bibliography{ref}
\bibliographystyle{icml2023}

\newpage
\appendix
\onecolumn

\input{sections/appendix}

\end{document}

%% file: defs.tex
\newcommand{\img}{\mathbf{x}}
\newcommand{\txt}{\mathbf{z}}

\DeclareMathOperator*{\expec}{\mathbb{E}}

%% file: sections/intro.tex
\section{Introduction} \label{sec:intro}


Deep generative models have recently shown remarkable success in generating high-quality images from text prompts~\cite{dalle1,dalle2,imagen,parti,stablediffusion}.
This success has been driven in part by the scaling of deep generative models 
to large-scale datasets from the web such as LAION~\cite{laion400m,laion-5b}.
However, major challenges remain in domains where large-scale text-to-image models fail to generate images that are well-aligned with text prompts~\cite{feng2022training,liu2022compositional,liu2022character}.
For instance, current text-to-image models often fail to produce reliable visual text~\cite{liu2022character} and struggle with \emph{compositional} image generation~\cite{feng2022training}.


In language modeling, \emph{learning from human feedback} has emerged as a powerful solution for aligning model behavior with human intent~\cite{ziegler2019fine,stiennon2020learning,wu2021recursively,webgpt,instructGPT,bai2022training}. Such methods first learn a \emph{reward function} intended to reflect what humans care about in the task, using human feedback on model outputs. The language model is then optimized using the learned reward function by a \emph{reinforcement learning (RL)} algorithm, such as proximal policy optimization~(PPO; \citealt{ppo}). This \emph{RL with human feedback (RLHF)} framework has successfully aligned large-scale language models (e.g., GPT-3;~\citealt{gpt3}) with complex human
quality assessments.

\begin{figure*} [t] \centering
\includegraphics[width=.95\textwidth]{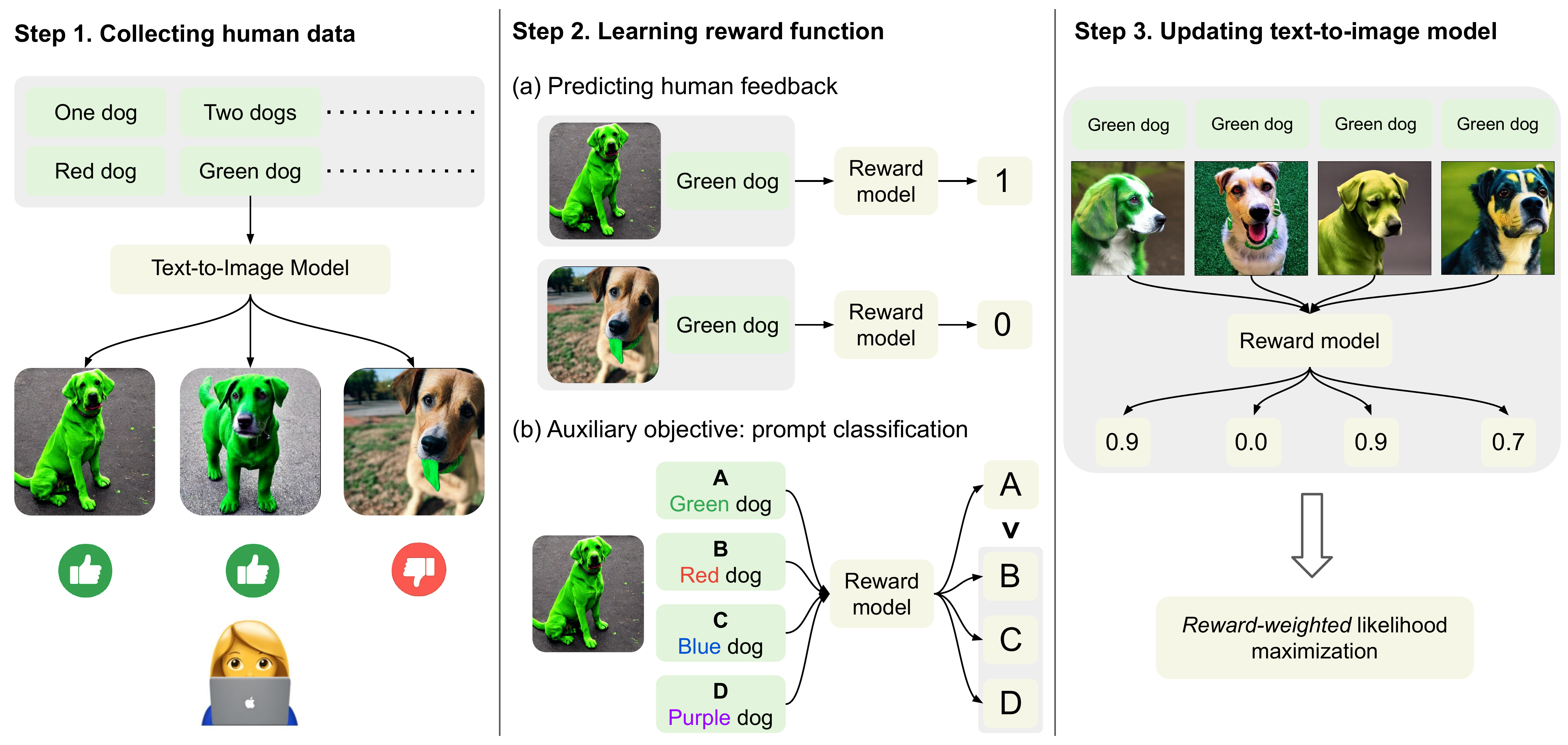}
\caption{The steps in our fine-tuning method. (1) Multiple images
sampled from
the text-to-image model using the same text prompt, followed by collection of (binary) human feedback. (2) A reward function is learned from human assessments to predict image-text alignment. We also utilize an auxiliary objective called prompt classification, which identifies the original text prompt within a set of \emph{perturbed} text prompts. (3) We update the text-to-image model via reward-weighted likelihood maximization.}
\label{fig:framework}
\end{figure*}

Motivated by the success of 
RLHF
in language domains, we propose a fine-tuning method for aligning text-to-image models using human feedback.
Our method consists of the following steps illustrated in Figure~\ref{fig:framework}: 
{\bf (1)} We first generate diverse images from a set of text prompts designed to test output alignment of a text-to-image model. Specifically, we examine prompts where pre-trained models are more prone to errors -- generating objects with specific colors, counts, and backgrounds. We then collect binary human feedback assessing model outputs.
{\bf (2)} Using this human-labeled dataset, we train a reward function to predict human feedback given the image and text prompt. We propose an auxiliary task---identifying the original text prompt within a set of \emph{perturbed} text prompts---to more effectively exploit human feedback for reward learning. 
This technique improves the generalization of reward function to unseen images and text prompts.
{\bf (3)} We update the text-to-image model via reward-weighted likelihood maximization to better align it with human feedback. Unlike the prior work~\cite{stiennon2020learning,instructGPT} that uses RL for optimization, we update the model using semi-supervised learning to measure model-output quality w.r.t.\ the learned reward function. 

We fine-tune the \emph{stable diffusion} model~\cite{stablediffusion} using 27K image-text pairs with human feedback. Our fine-tuned model shows improvement in generating objects with specified colors, counts, and backgrounds. Moreover, it improves compositional generation (i.e., can better generate unseen objects\footnote{We use the term ``unseen objects'' w.r.t. our human dataset, i.e., they are not included in the human dataset but pre-training dataset would contain them.} given unseen combinations of color, count, and background prompts). We also observe that the learned reward function is better aligned with human assessments of alignment than CLIP score~\cite{clip} on tested text prompts. We analyze several design choices, such as using an auxiliary loss for reward learning and the effect of using ``diverse'' datasets for fine-tuning.


We can summarize our main contributions as follows:
\begin{itemize} [leftmargin=4mm]
\setlength\itemsep{0.1em}
\item We propose a simple yet efficient fine-tuning method for aligning a text-to-image model using human feedback.
\item We show that fine-tuning with human feedback significantly improves the image-text alignment of a text-to-image model. On human evaluation, our model achieves up to 47\% improvement in image-text alignment at the expense of mildly degraded image fidelity.
\item We show that the learned reward function predicts human assessments of the quality more accurately than the CLIP score~\cite{clip}. In addition, we show that rejection sampling based on our learned reward function can also significantly improve the image-text alignment.
\item Naive fine-tuning with human feedback can significantly reduce the image fidelity, despite better alignment. We find that careful investigations on several design choices are important in balancing alignment-fidelity tradeoffs. 
\end{itemize}

Even though our results do not address all the failure modes of the existing text-to-image models, we hope that this work highlights the potential of learning from human feedback for aligning these models.

%% file: sections/relatedwork.tex
\section{Related Work} \label{sec:related}



\paragraph{Text-to-image models.}
Various deep generative models, such as variational auto-encoders 
\cite{vae}, 
generative adversarial networks 
\cite{gan}, 
auto-regressive models~\cite{van2016pixel}, and diffusion models~\cite{sohl2015deep,ho2020denoising} have been proposed for image distributions.
Combined with the large-scale language encoders~\cite{clip,t5}, these models have shown impressive results in text-to-image generation~\cite{dalle2,imagen,parti,stablediffusion}.

However, text-to-image models frequently struggle to generate images that are well-aligned with text prompts~\cite{feng2022training,liu2022compositional,liu2022character}. 
\citet{liu2022character} show that current models fail to produce reliable visual text and often perform poorly w.r.t.\ compositional generation~\cite{feng2022training,liu2022compositional}.
Several techniques, such as character-aware text encoders~\cite{xue2022byt5,liu2022character} and structured representations of language inputs~\cite{feng2022training} have been investigated to address these issues.
We study learning from human feedback, which aligns text-to-image models directly using human feedback on model outputs. 

Fine-tuning with few images~\cite{dreambooth,kumari2022multi,gal2022image} for personalization of text-to-image diffusion models is also related with our work.
DreamBooth~\cite{dreambooth} showed that text-to-image models can generate diverse images in a personalized way by fine-tuning with few images, and \citet{kumari2022multi} proposed a more memory and computationally efficient method.
In this work, we demonstrate that it is possible to fine-tune text-to-image models using simple binary (good/bad) human feedback.

\vspace{-0.06in}
\paragraph{Learning with human feedback.}
Human feedback has been used to improve various AI systems, from translation~\citep{bahdanau2016actor, kreutzer2018can}, to web question-answering~\citep{webgpt}, to story generation~\citep{zhou2020learning}, to training RL agents without the need for hand-designed rewards~\citep[][inter alia]{coach,preference_drl,deeptamer,ibarz2018preference_demo,pebble}, and to more truthful and harmless instruction following and dialogue~\citep[][inter alia]{instructGPT, bai2022training, ziegler2019fine,wu2021recursively,stiennon2020learning, liu2023chain, scheurer2022training, bai2022constitutional}. 
In relation to prior works that focus on improving language models and game agents with human feedback, our work explores using human feedback to align multi-modal text-to-image models with human preference.
Many prior works on learning with human feedback consist of learning a reward function and maximizing reward weighted likelihood (often dubbed as supervised fine-tuning)~\citep[see e.g.][]{instructGPT, ziegler2019fine, stiennon2020learning}
Inspired by their successes, we propose a fine-tuning method with human feedback for improving text-to-image models.

\vspace{-0.06in}
\paragraph{Evaluating image-text alignment.}
To measure the image-text alignment, various evaluation protocols have been proposed~\cite{madhyastha2019vifidel,hessel2021clipscore,imagen,parti}.
Most prior works~\cite{dalle2,imagen,parti} use the alignment score of image and text embeddings determined by pre-trained multi-modal models, such as CLIP~\cite{clip} and CoCa~\cite{yu2022coca}. However, since scores from pre-trained models are not calibrated with human intent, human evaluation has also been introduced~\cite{imagen,parti}. In this work, we train a reward function that is better aligned with human evaluations by exploiting pre-trained representations and (a small amount) of human feedback data.

%% file: sections/method.tex
\section{Main Method} \label{sec:method}

To improve the alignment of generated images with their text prompts, 
we fine-tune a pre-trained text-to-image model~\cite{dalle2,imagen,stablediffusion} by repeating the following steps shown in Figure~\ref{fig:framework}. We first generate a set of diverse images from a collection of text prompts designed to test various capabilities of the text-to-image model. Human raters provide binary feedback
on these images (Section~\ref{sec:human}).
Next we train a \emph{reward model} to predict human feedback given a text prompt and an image as inputs (Section~\ref{sec:reward}).
Finally, we fine-tune the text-to-image model using \emph{reward-weighted} log likelihood to improve text-image alignment (Section~\ref{sec:finetune}).


\subsection{Human Data Collection} \label{sec:human}

\paragraph{Image-text dataset.} 
To test specific capabilities of a given text-to-image model, we consider three categories of text prompts that generate objects with a specified count, color, or background.\footnote{For simplicity, we consider a limited class of text categories in this work, deferring the study of broader and more complex categories to future work.}
For each category, we generate prompts by combining a word or phrase from that category with some object; e.g., combining {\tt green} (or {\tt in a city}) with {\tt dog}. 
We also consider combinations of the three categories (e.g., {\tt two green dogs in a city}).
From each prompt, we generate up to 60 images using a pre-trained text-to-image model---in this work, we use Stable Diffusion v1.5~\cite{stablediffusion}.

\paragraph{Human feedback.} 
We collect simple binary feedback from multiple human labelers on the image-text dataset. Labelers are presented with three images generated from the same prompt and are asked to assess whether each image is well-aligned with the prompt (``good'') or not (``bad'').\footnote{Labelers are instructed to skip a query if it is hard to answer. Skipped queries are not used in training.}
We use binary feedback given the simplicity of our prompts---the evaluation criterion are fairly clear. More informative human feedback, such as ranking~\cite{stiennon2020learning,instructGPT}, should prove useful when more complex or subjective text prompts are used (e.g., artistic or open-ended generation).



\subsection{Reward Learning} \label{sec:reward}

To measure image-text alignment, we learn a \emph{reward function} $r_\phi (\img,\txt)$ (parameterized by $\phi$) that
maps the CLIP embeddings\footnote{To improve generalization ability, we use CLIP embeddings pre-trained on various image-text samples.}~\cite{clip} of an image $\img$ and a text prompt $\txt$ to a scalar value. It is trained to 
predict human feedback $y\in\{0,1\}$ (1 = good, 0 = bad). 

Formally, given the human feedback dataset $\mathcal{D}^{\tt human}=\{\left(\img,\txt,y\right)\}$, 
the reward function $r_\phi$
is trained by minimizing the mean-squared-error (MSE):
\begin{align*}
    \mathcal{L}^{\tt MSE} (\phi) = \expec_{\left(\img,\txt,y\right) \sim \mathcal{D}^{\tt human}} \big[\left(y - r_\phi (\img,\txt) \right)^2\big].
\end{align*}



\paragraph{Prompt classification.}
Data augmentation can significantly improve the data-efficiency and performance of learning~\cite{krizhevsky2017imagenet,cubuk2019autoaugment}.
To effectively exploit the feedback dataset, we design a simple data augmentation scheme and auxiliary loss for reward learning. For each image-text pair that has been labeled {\em good}, we generate $N-1$ text prompts with different semantics than the original text prompt. For example, we might generate \{{\tt Blue dog} , $\ldots$ , {\tt Green dog}\} given the original prompt {\tt Red dog}.\footnote{We use a rule-based strategy to generate different text prompts (see Appendix~\ref{app:pseudo} for more details).} This process generates a dataset $\mathcal{D}^{\tt txt} = \{(\img, \{\txt_j\}_{j=1}^N, i^\prime)\}$ with $N$ text prompts $\{\txt_j\}_{j=1}^N$, including the original, for each image $\img$, and the index $i'$ of the original prompt.

We use the augmented prompts in an auxiliary task, namely, classifying the original prompt for reward learning. Our prompt classifier uses the reward function $r_\phi$ as follows:
\begin{align*}
    P_\phi ( i| \img, \{\txt_j\}_{j=1}^N) = \frac{\exp(r_\phi (\img,\txt_i) / T)}{\sum_j \exp(r_\phi (\img,\txt_j)/ T)},\quad \forall i\in[N],
\end{align*}
where $T>0$ is the temperature.
Our auxiliary loss is
\begin{align}
    \mathcal{L}^{\tt pc}(\phi) = \!\!\!\!\!\!\! \expec_{(\img, \{\txt_j\}_{j=1}^N, i^\prime)\sim \mathcal{D}^{\tt txt}} \!\!\big[  \mathcal{L}^{\tt CE} \big(P_\phi ( i| \img, \{\txt_j\}_{j=1}^N ), i^\prime\big) \big],\!\!\! \label{eq:pc}
\end{align}
where $\mathcal{L}^{\tt CE}$ is the standard cross-entropy loss. 
This encourages $r_\phi$ to produce low values for prompts with different semantics than the original. Our experiments show this auxiliary loss improves the generalization to unseen images and text prompts. Finally, we define the combined loss as
\begin{align*}
    \mathcal{L}^{\tt reward} (\phi) = \mathcal{L}^{\tt MSE}(\phi) + \lambda  \mathcal{L}^{\tt pc}(\phi),
\end{align*}
where $\lambda$ is the penalty parameter. 

The pseudo code of reward learning is in Appendix~\ref{app:pseudo}.



\subsection{Updating the Text-to-Image Model} \label{sec:finetune}

We use our learned $r_\phi$ to update the text-to-image model $p$ with parameters $\theta$ by minimizing the loss
\begin{equation}
\label{eq:text-to-image-loss}
\begin{split}
    \mathcal{L} (\theta) &= \expec_{(\img,\txt)\sim\mathcal{D}^{\tt model}} \big[- r_\phi (\img,\txt) \log p_\theta(\img|\txt)\big] \\
    &\qquad\quad + \beta \expec_{(\img,\txt)\sim\mathcal{D}^{\tt pre}}\big[-\log p_\theta(\img|\txt) \big], 
\end{split}
\end{equation}
where $\mathcal{D}^{\tt model}$ is the model-generated dataset (i.e., images generated by the text-to-image model on the tested text prompts),
$\mathcal{D}^{\tt pre}$ is the \emph{pre-training dataset}, and $\beta$ is a penalty parameter. The first term in~\eqref{eq:text-to-image-loss} minimizes the \emph{reward-weighted} negative log-likelihood (NLL) on
$\mathcal{D}^{\tt model}$.\footnote{To increase diversity, we collect an unlabeled dataset $\mathcal{D}^{\tt unlabel}$ by generating more images from the text-to-image model, and use both the human-labeled dataset $\mathcal{D}^{\tt human}$ and the unlabeled dataset $\mathcal{D}^{\tt unlabel}$ for training, i.e.,~$\mathcal{D}^{\tt model} = \mathcal{D}^{\tt human} \cup \mathcal{D}^{\tt unlabel}$.}
By evaluating the quality of the outputs using a reward function aligned with the text prompts, this term improves the image-text alignment of the model.



Typically, the diversity of the model-generated dataset is limited, which can result in overfitting. To mitigate this, similar to~\citet{instructGPT}, we also minimize the \emph{pre-training loss}, the second term in~\eqref{eq:text-to-image-loss}. This reduces NLL on the pre-training dataset $\mathcal{D}^{\tt pre}$. In our experiments, we observed regularization in the loss function $\mathcal{L} (\theta)$ in~\eqref{eq:text-to-image-loss} enables the model to generate more natural images.


Different objective functions and algorithms (e.g., PPO;~\citealt{ppo}) could be considered for updating the text-to-image model similar to RLHF fine-tuning~\cite{instructGPT}. We believe RLHF fine-tuning may lead to better models because it uses online sample generation during updates and KL-regularization over the prior model. However, RL usually requires extensive hyperparameter tuning and engineering, thus, we defer the extension to RLHF fine-tuning to future work.

%% file: sections/experiments.tex
\begin{table}[t!]
\begin{center}
 \small
\begin{tabular}{lc}
\toprule
Category & Examples \\ \midrule
\multirow{2}{*}{\begin{tabular}[c]{@{}c@{}} {Count} \end{tabular}} 
& {\tt One dog}; $\;$ {\tt Two dogs}; $\;$ {\tt Three dogs}; \\ 
& {\tt Four dogs}; $\;$ {\tt Five dogs}; \\  \midrule
\multirow{2}{*}{\begin{tabular}[c]{@{}c@{}} {Color} \end{tabular}} 
&  {\tt A green colored dog};  \\
&{\tt A red colored dog};  \\  \midrule
\multirow{2}{*}{\begin{tabular}[c]{@{}c@{}} {Background} \end{tabular}} 
& {\tt A dog in the forest};\\
& {\tt A dog on the moon};    \\ \midrule
\multirow{2}{*}{\begin{tabular}[c]{@{}c@{}} {Combination} \end{tabular}} 
&  {\tt Two blue dogs in the forest};  \\ 
& {\tt Five white dogs in the city}; \\
\bottomrule
\end{tabular}
\end{center}
\caption{Examples of text categories.}
\label{table:text_example}
\end{table}

\begin{table}[t!]
\begin{center}
 \small
\begin{tabular}{lclll}
\toprule
\multirow{2}{*}{\begin{tabular}[c]{@{}c@{}} {Category} \end{tabular}} 
& \multirow{2}{*}{\begin{tabular}[c]{@{}c@{}} {Total \# of} \\ {images} \end{tabular}} 
& \multicolumn{3}{c}{Human feedback (\%)} \\ \cmidrule(lr){3-5}
& &  {Good} & {Bad} & {Skip} \\
\midrule
Count & $6480$ & $34.4$ & $61.0$ & $4.6$ \\
Color & $3480$  & $70.4$ & $20.8$ & $8.8$   \\
Background & $2400$  & $66.9$ & $33.1$ & $0.0$   \\
Combination & $15168$  & $35.8$ & $59.9$ & $4.3$   \\  \midrule
Total & $27528$ & $46.5$ &$48.5$ &$5.0$  \\
\bottomrule
\end{tabular}
\end{center}
\caption{Details of image-text datasets and human feedback.}
\label{table:text_dist}
\end{table}

\begin{figure*} [t!] \centering
\subfigure[Seen text prompt: {\tt Two green dogs on the table.}]
{
\includegraphics[width=0.99\textwidth]{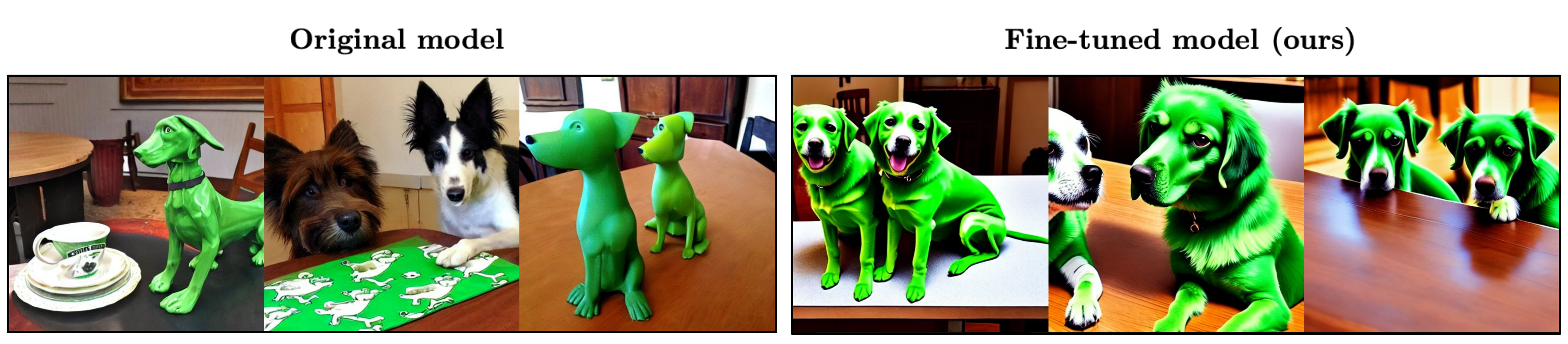}
\label{fig:main_fig_0}} 
\subfigure[Unseen text prompt (unseen object): {\tt Four tigers in the field.}]
{\includegraphics[width=0.99\textwidth]{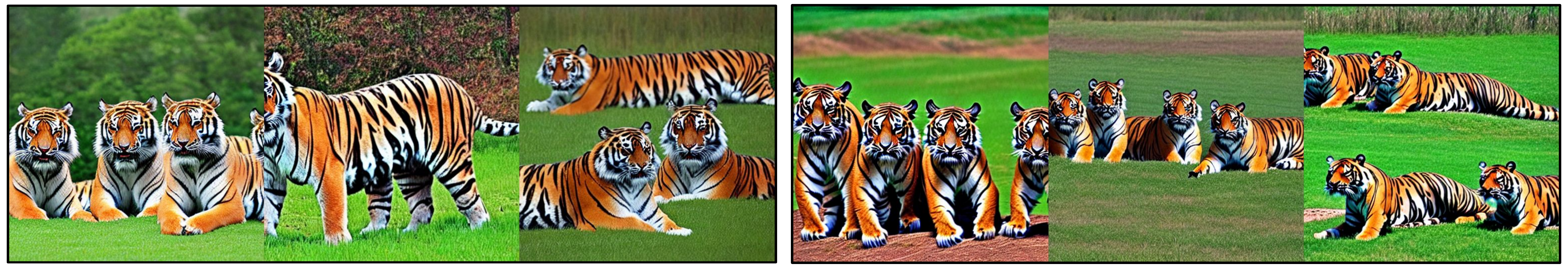}
\label{fig:main_fig_2}}
\subfigure[Unseen text prompt (artistic generation): {\tt Oil painting of sunflowers.}]
{\includegraphics[width=0.99\textwidth]{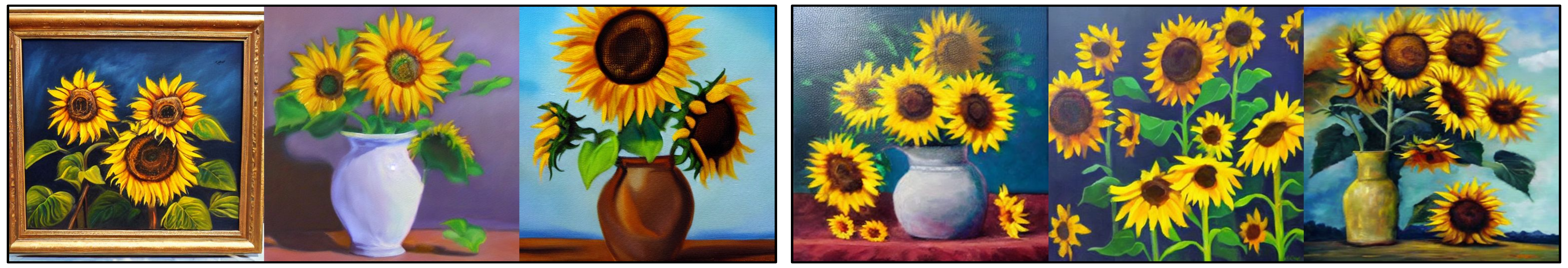}
\label{fig:main_fig_3}}
\caption{Samples from the original Stable Diffusion model (left) and our fine-tuned model (right). (a) Our model generates high-quality seen object ({\tt dog}) with specified color, count and background. on seen text prompts. 
(b) Our model generates an unseen object ({\tt tiger}) with specified color, count, and background. (c) Our model still generates reasonable images from unseen text categories (artistic generation).}
\label{fig:main_fig}
\end{figure*}

\section{Experiments} \label{sec:exp}

We describe a set of experiments designed to test the efficacy of our fine-tuning approach with human feedback.

\subsection{Experimental Setup}

\paragraph{Models.}
For our baseline generative model, 
we use stable diffusion v1.5~\cite{stablediffusion}, which has been pre-trained on large image-text datasets~\cite{laion400m,laion-5b}.\footnote{Our fine-tuning method can be used readily with other text-to-image models, such as Imagen~\cite{imagen}, Parti~\cite{parti} and Dalle-2~\cite{dalle2}.} For fine-tuning, we freeze the CLIP language encoder~\cite{clip} and fine-tune only the diffusion module.
For the reward model, we use ViT-L/14 CLIP model~\cite{clip} to extract image and text embeddings and train a MLP using these embeddings as input. More experimental details (e.g., model architectures and the final hyperparameters) are reported in Appendix~\ref{app:exp_detail}.

\paragraph{Datasets.} 
From a set of 2700 English prompts (see Table~\ref{table:text_example} for examples), we generate 27K images using the stable diffusion model (see Appendix~\ref{app:dataset} for further details).
Table~\ref{table:text_dist} shows the feedback distribution provided by multiple human labelers, which has been class-balanced. We note that the stable diffusion model struggles to generate the number of objects specified by the prompt, but reliably generates specified colors and backgrounds.

We use 23K samples for training, with the remaining samples used for validation. 
We also use 16K unlabeled samples for the reward-weighted loss
and a 625K subset\footnote{\url{https://huggingface.co/datasets/ChristophSchuhmann/improved_aesthetics_6.5plus}} of LAION-5B~\cite{laion-5b} filtered by an aesthetic score predictor \footnote{\url{https://github.com/christophschuhmann/improved-aesthetic-predictor}} for the pre-training loss.



\begin{figure*} [t!] \centering
\subfigure[Comparison with CLIP score~\cite{clip}]
{\includegraphics[width=0.43\textwidth]{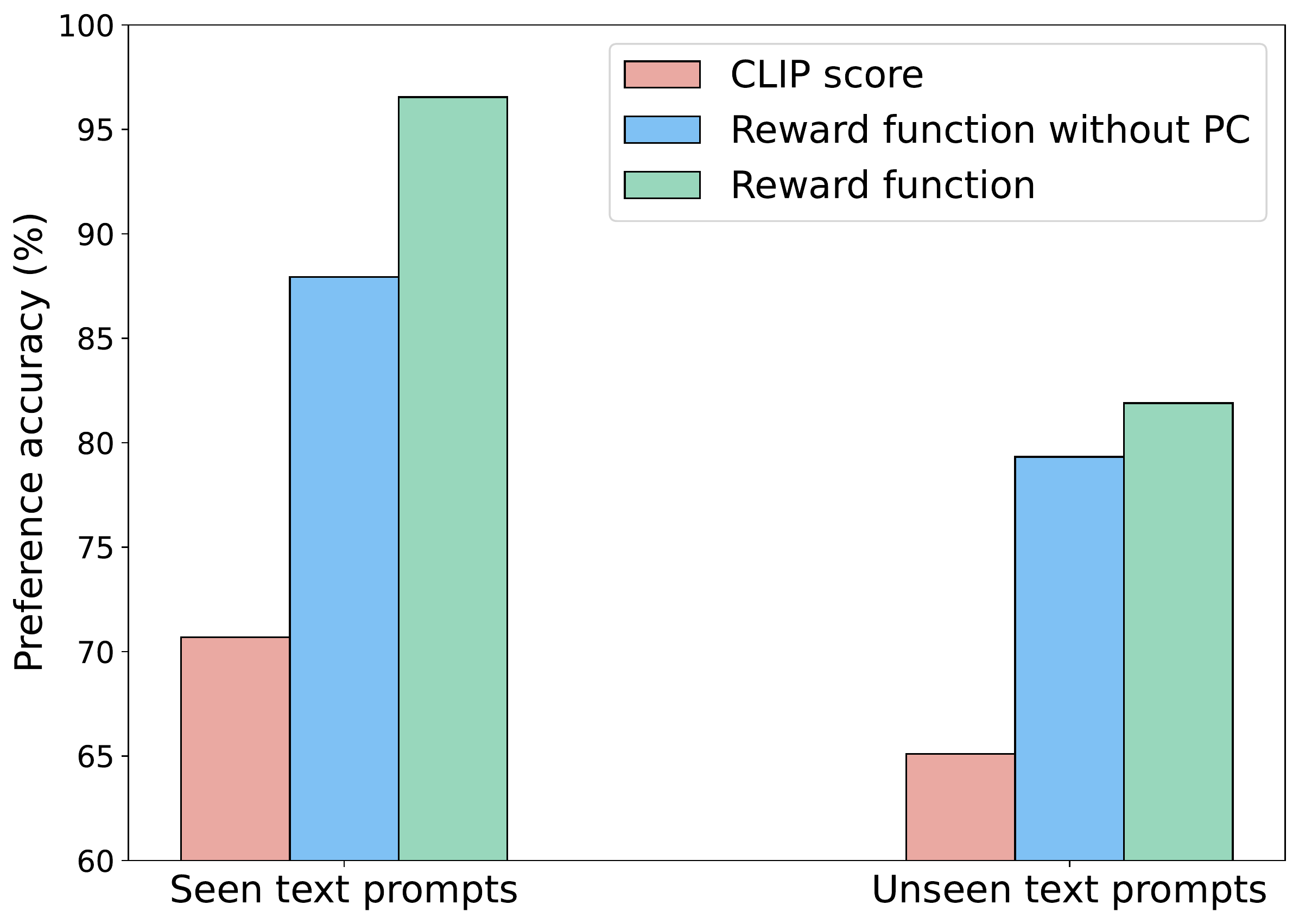}
\label{fig:comp_clip}} 
\subfigure[Performance of reward function versus data size]
{\includegraphics[width=0.43\textwidth]{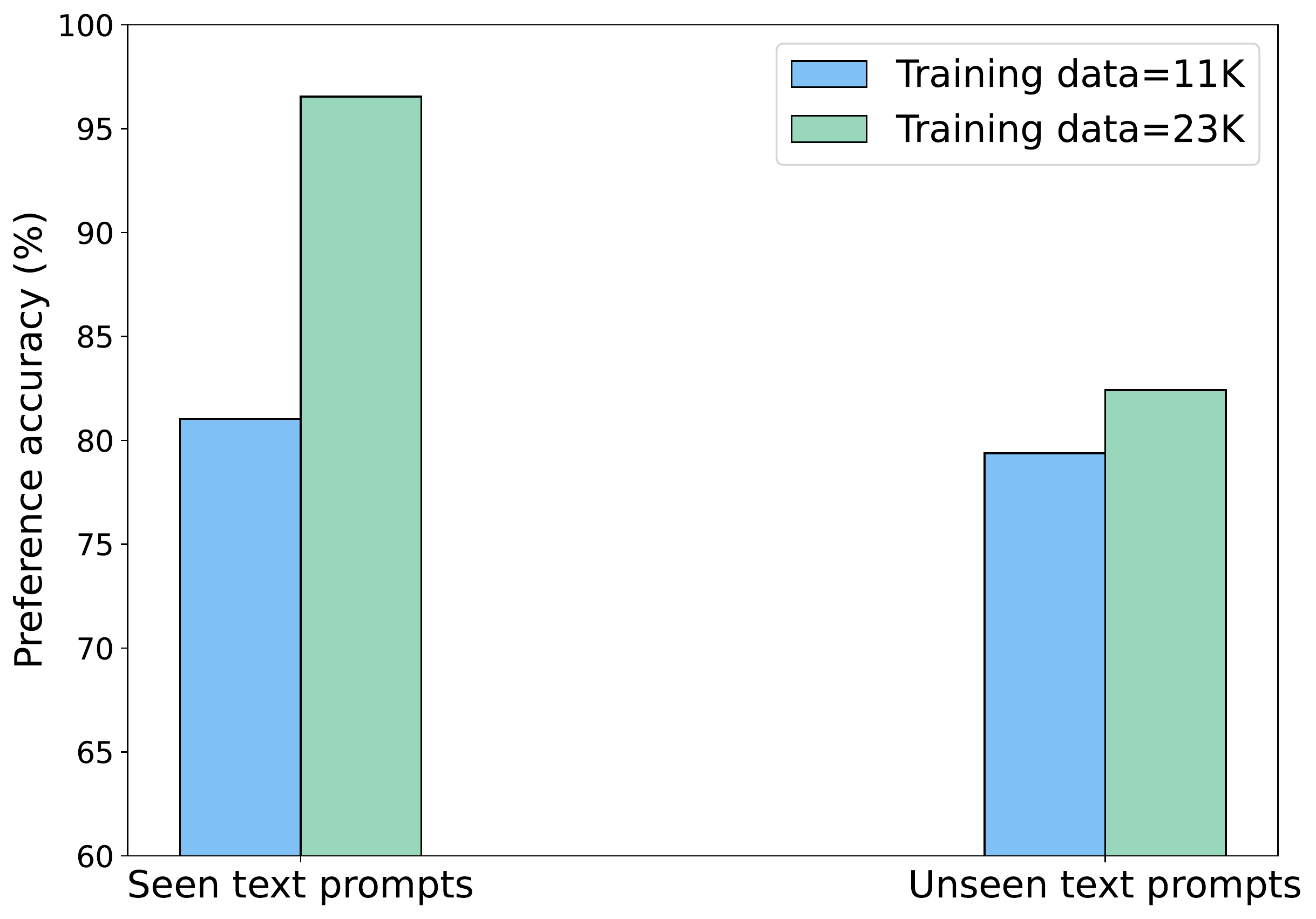}
\label{fig:human_size}}
\caption{(a) Accuracy of CLIP score~\cite{clip} and our reward functions on predicting the preferences of human labelers. For our method, we consider a variant of our reward function, which is not trained with prompt classification (PC) loss in \eqref{eq:pc}. (b) Performance of reward functions with varying the size of the training dataset.}
\label{fig:reward}
\end{figure*}

\begin{figure} [t] \centering
\includegraphics[width=.5\textwidth]{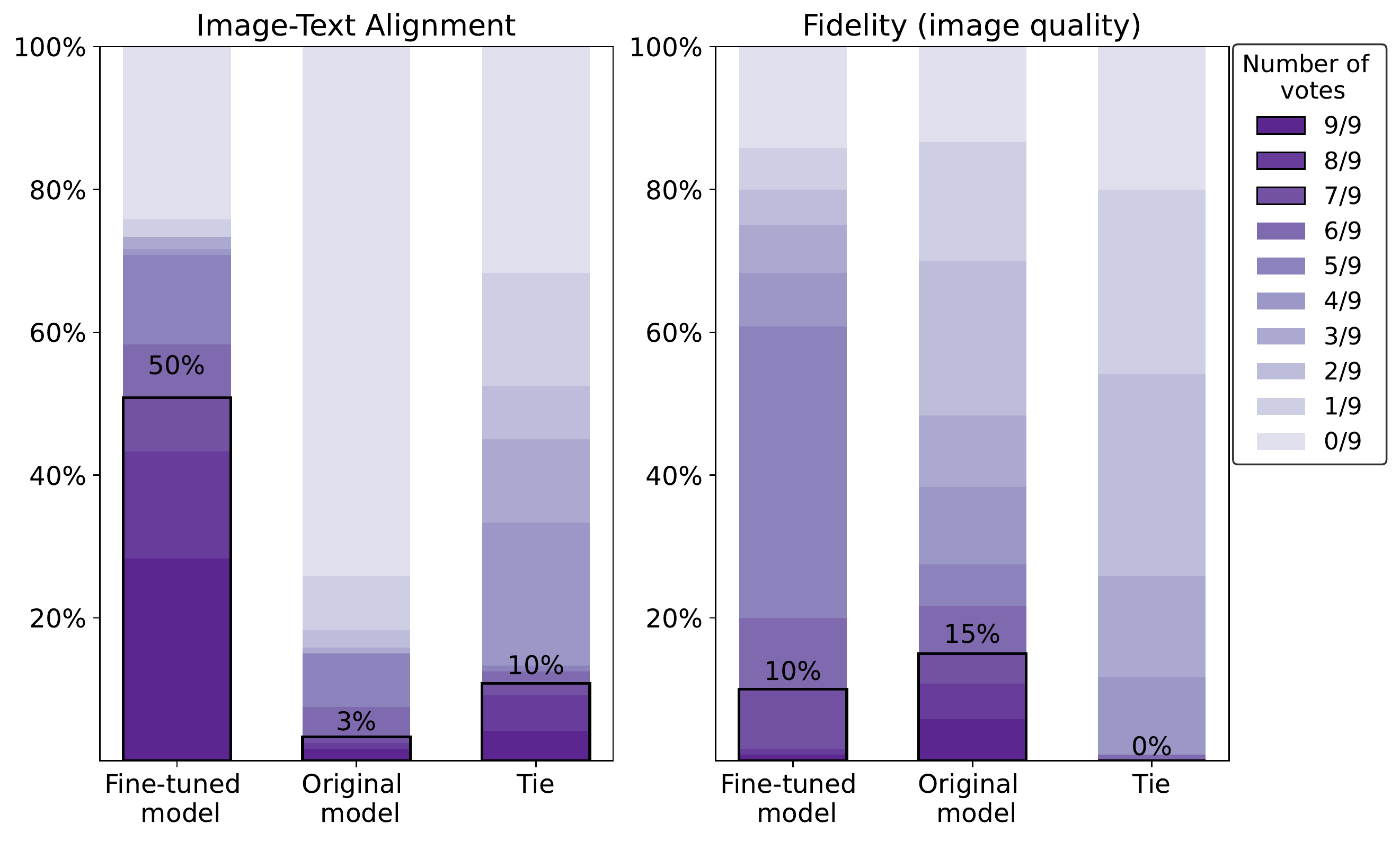}
\caption{Human evaluation results on 120 text prompts (60 seen text prompts and 60 unseen text prompts). We generate two sets of images (one from our fine-tuned model and one from the original model) with same text prompt. Then, human raters indicate which one is better, or tie (i.e., two sets are similar) in terms of image-text alignment and image fidelity. Each query is evaluated by 9 independent human raters and we report the percentage of queries based on the number of positive votes. We also highlight the percentage of queries with two-thirds vote (7 or more positive votes) in the black box.}
\label{fig:main}
\end{figure}


\begin{figure*} [t!] \centering
\subfigure[Fine-tuned model only with human-labeled dataset.]
{
\includegraphics[width=0.99\textwidth]{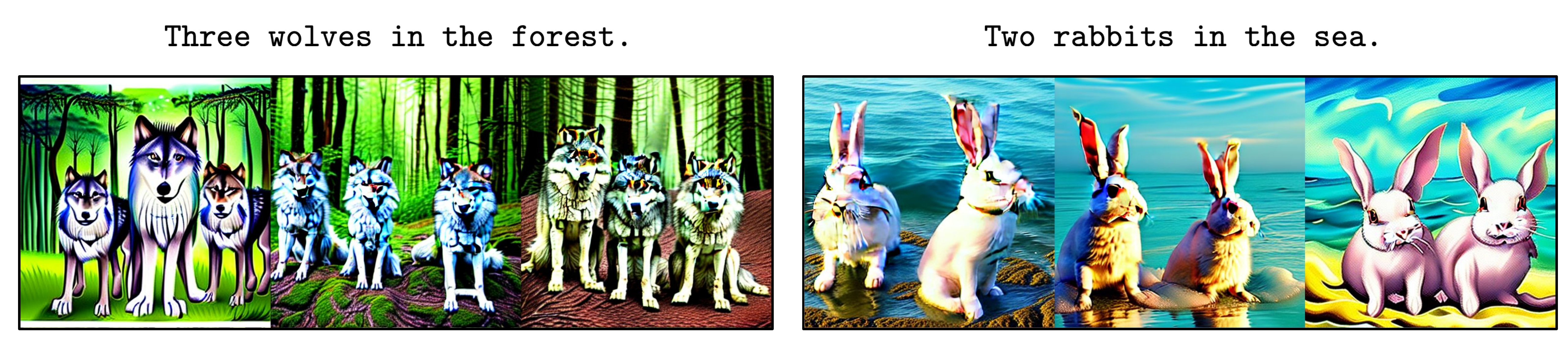}
\label{fig:abl_fig_0}} 
\subfigure[Fine-tuned model with human-labeled and unlabeled datasets.]
{\includegraphics[width=0.99\textwidth]{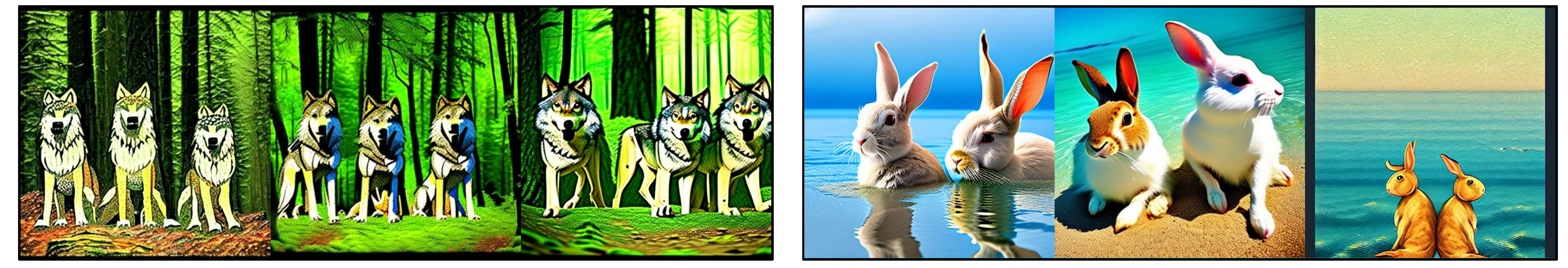}
\label{fig:abl_fig_1}}
\subfigure[Fine-tuned model with human-labeled, unlabeled and pre-training datasets.]
{\includegraphics[width=0.99\textwidth]{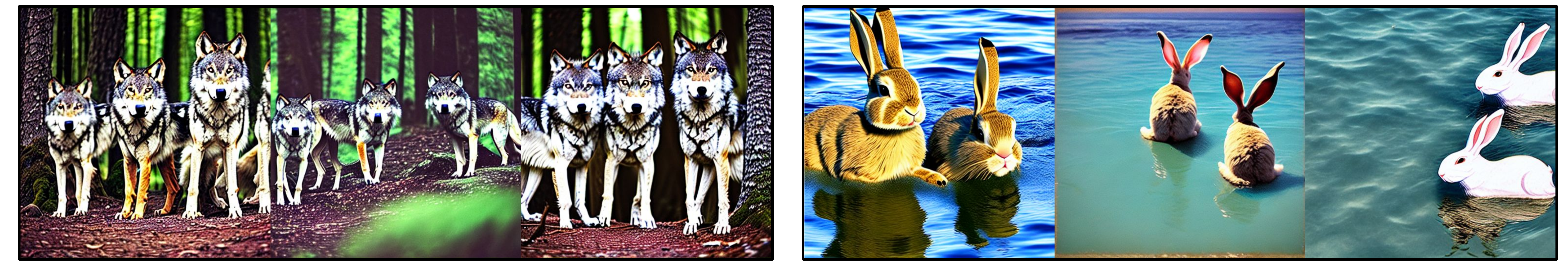}
\label{fig:abl_fig_2}}
\vspace{-0.1in}
\caption{Samples from fine-tuned models trained with different datasets on unseen text prompts. (a) Fine-tuned model only with human dataset generates low-quality images due to overfitting. (b) Unlabeled samples improve the quality of generated images. (c) Fine-tuned model can generate high-fidelity images by utilizing pre-training dataset.}
\label{fig:abl_fig}
\end{figure*}

\subsection{Text-Image Alignment Results} \label{sec:main_result}

\paragraph{Human evaluation.}
We measure human ratings of image alignment with 120 text prompts (60 seen text prompts and 60 unseen\footnote{Here, ``unseen'' text prompts consist of ``unseen'' objects, which are not in our human dataset.} text prompts), testing the ability of the models to render different colors, number of objects, and backgrounds (see Appendix~\ref{app:dataset} for the full set of prompts). 
Given two (anonymized) sets of images, one from our fine-tuned model and one from the stable diffusion model, we ask human raters to assess which is better w.r.t.\ image-text alignment and fidelity (i.e., image quality).\footnote{We ask raters to declare a {\em tie} if they have similar quality.} Each query is evaluated by 9 independent human raters. We show the percentage of queries based on the number of positive votes.

As shown in Figure~\ref{fig:main},
our method significantly improves image-text alignment against the original model. 
Specifically, 50\% of samples from our model receive at least two-thirds vote (7 or more positive votes) for image-text alignment.
However, fine-tuning somewhat degrades image fidelity (15\% compared to 10\%). 
We expect that this is because (i) we asked the labelers to provide feedback mainly on alignment, (ii) the diversity of our human data is limited, and (iii) we used a small subset of pre-training dataset for fine-tuning.\footnote{Similar issue, which is akin to the {\em alignment tax}, has been observed in language domains~\cite{,askell2021general,instructGPT}.} This issue can presumably be mitigated with larger rater and pre-training datasets. 


\paragraph{Qualitative comparison.} 
Figure~\ref{fig:main_fig} shows image samples from the original model and our fine-tuned counterpart (see Appendix~\ref{app:example} for more image examples). While the original often generates images with missing details (e.g., color, background or count) (Figure~\ref{fig:main_fig_0}), our model generates objects that adhere to the prompt-specified colors, counts and backgrounds.
Of special note, our model generates high-quality images on unseen text prompts that specify unseen objects (Figure~\ref{fig:main_fig_2}).
Our model also generates reasonable images given unseen text categories, such as artistic generation (Figure~\ref{fig:main_fig_3}).

However, we also observe several issues of our fine-tuned models.
First, for some specific text prompts, 
our fine-tuned model generates oversaturated and non-photorealistic images. 
Our model occasionally duplicates entities within the generated images or produces lower-diversity images for the same prompt.
We expect that it would be possible to address these issues with larger (and diverse) human datasets and better optimization (e.g., RL). 



\subsection{Results on Reward Learning}

\paragraph{Predicting human preferences.} 
We investigate the quality of our learned reward function by evaluating its prediction of with human ratings. Given two images from the same text prompt $(\img_1,\img_2,\txt)$, we check whether our reward $r_\phi$ generates a higher score for the human-preferred image, i.e., $r_\phi(\img_1, \txt) > r_\phi(\img_2, \txt)$ when rater prefers $\img_1$. As a baseline, we compare it with the CLIP score~\cite{hessel2021clipscore}, which measures image-text similarity in the CLIP embedding space~\cite{clip}.

Figure~\ref{fig:comp_clip} compares the accuracy of $r_\phi$ and the CLIP score on unseen images from both seen and unseen text prompts. Our reward (green) more accurately predicts human evaluation than the CLIP score (red), hence is better aligned with typical human intent.
To show the benefit of our auxiliary loss (prompt classification) in \eqref{eq:pc}, we also assess a variant of our reward function which ignores the auxiliary loss (blue). 
The auxiliary classification task improves reward performance on both seen and unseen text prompts.
The gain from the auxiliary loss clearly shows the importance of text diversity and our auxiliary loss in improving data efficiency.
Although our reward function is more accurate than the CLIP score, its performance on unseen text prompts ($\sim 80\%$) suggests that it may be necessary to use more diverse and large human datasets.


\begin{figure*} [t!] \centering
\subfigure[Original model versus Rejection sampling]
{
\includegraphics[width=0.48\textwidth]{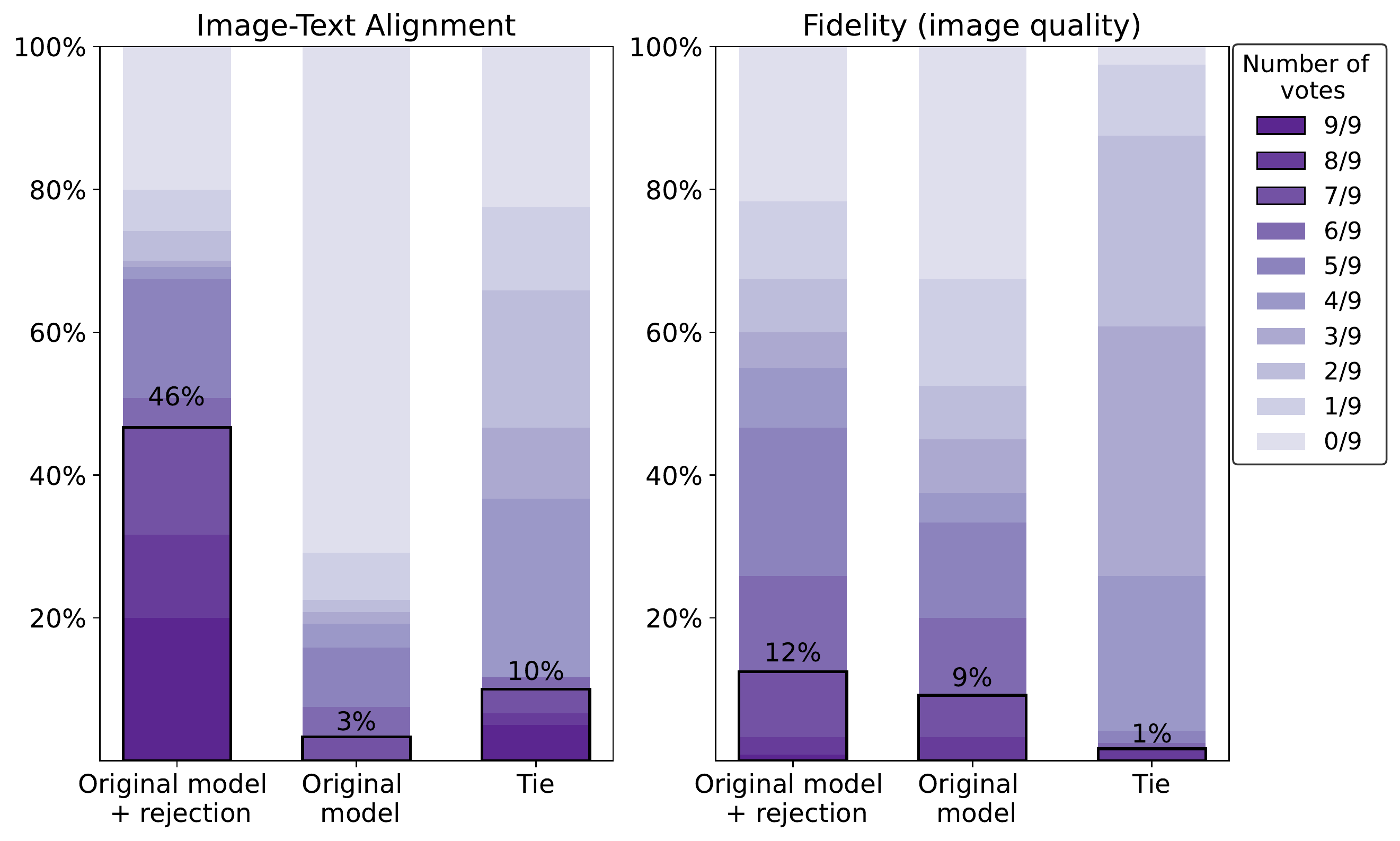}
\label{fig:rejection_1}} 
\subfigure[Fine-tuned model versus Rejection sampling]
{\includegraphics[width=0.48\textwidth]{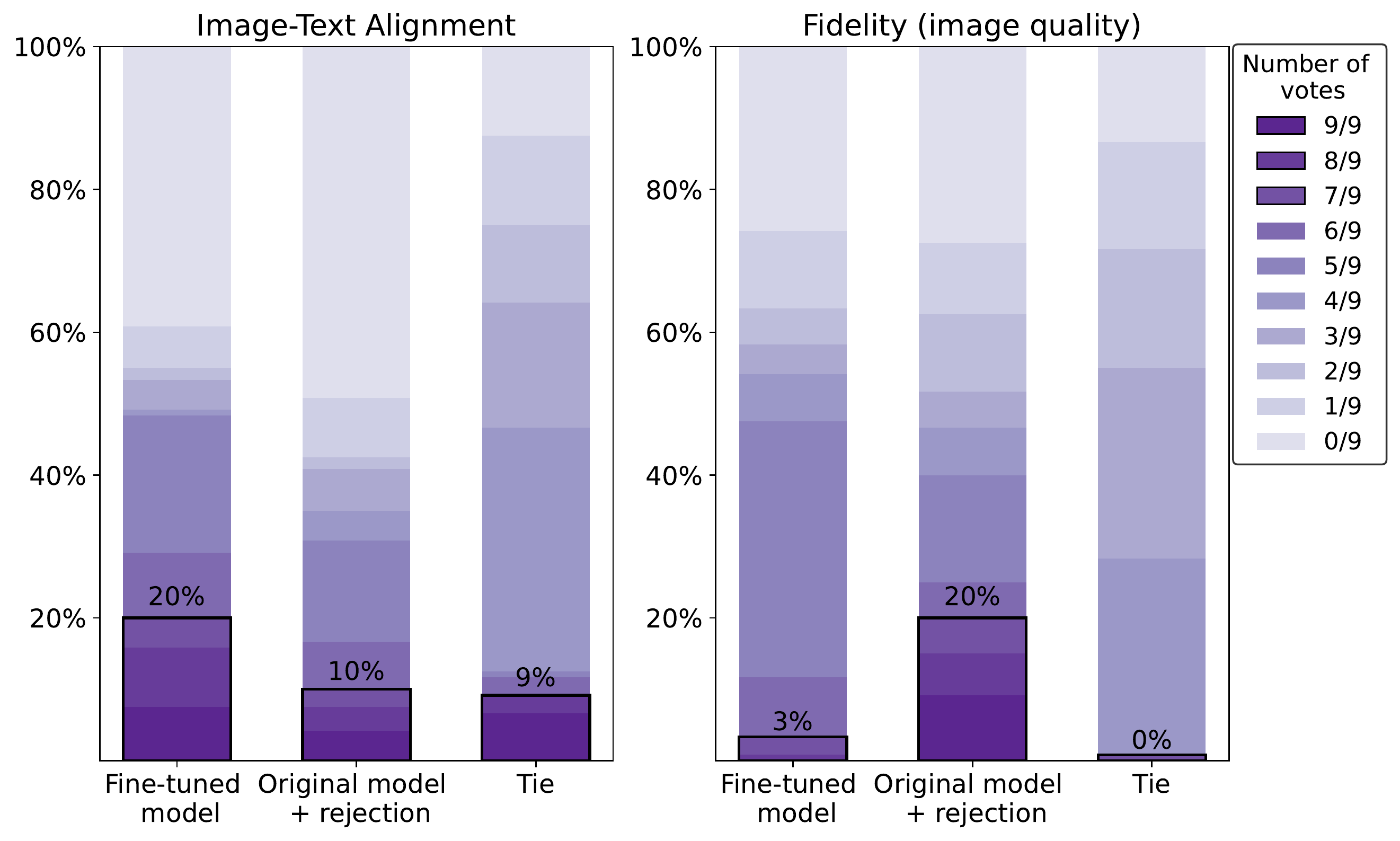}
\label{fig:rejection_2}}
\caption{Human evaluation on 120 tested text prompts (60 seen text prompts and 60 unseen text prompts). 
We generate two sets of images with same text prompt. Then, human raters indicate which one is better, or tie (i.e., two sets are similar) in terms for image-text alignment and image fidelity. Each query is evaluated by 9 independent human raters and we report the percentage of queries based on the number of positive votes. We also highlight the percentage of queries with two-thirds vote (7 or more positive votes) in the black box.
(a) For rejection sampling, we generate 16 images per text prompt and select best 4 images based on reward score, i.e., more inference-time compute. (b) Comparison between fine-tuned model and original model with rejection sampling. }
\label{fig:rejection}
\end{figure*}

\paragraph{Rejection sampling.} 
Similar to Parti~\cite{parti} and DALL-E~\cite{dalle1}, we evaluate a rejection sampling technique, which selects the best output w.r.t.\ the learned reward function.\footnote{Parti and DALL-E use similarity scores of image and text embeddings from CoCa~\cite{yu2022coca} and CLIP~\cite{clip}, respectively.}
Specifically, we generate 16 images per text prompt from the original stable diffusion model and select the 4 with the highest reward scores. We compare these to 4 randomly sampled images in Figure~\ref{fig:rejection_1}. 
Rejection sampling significantly improves image-text alignment (46\% with two-thirds preference vote by raters) without sacrificing image fidelity. This result illustrates the significant value of the reward function in improving text-to-image models \emph{without any fine-tuning}. 

We also compare our fine-tuned model to the original with rejection sampling in Figure~\ref{fig:rejection_2}.\footnote{We remark that rejection sampling can also be applied on top of our fine-tuned model.} 
Our fine-tuned model achieves a 10\% gain in image-text alignment (20\%-10\% two-thirds vote) but sacrifices 17\% in image fidelity (3\%-20\% two-thirds vote).
However, as discussed in Section~\ref{sec:main_result}, 
we expect degradation in fidelity to be mitigated with larger
human datasets and better hyper-parameters. 
Note also that rejection sampling has several drawbacks, including increased inference-time computation and the inability to improve the model (since it is output post-processing).


\subsection{Ablation Studies}

\paragraph{Effects of human dataset size.}
To investigate how human data quality affects reward learning, we conduct an ablation study,
reducing the number of images per text prompt by half before training the reward function. 
Figure~\ref{fig:human_size} shows that model accuracy decreases on both seen and unseen prompts as data size decreases, clearly demonstrating the importance of diversity and the amount of rater data.


\paragraph{Effects of using diverse datasets.}
To verify the importance of data diversity, we incrementally include unlabeled and pre-training datasets during fine-tuning.
We measure the reward score (image-text alignment) on 120 tested text prompts and FID score~\cite{fid}---the similarity between generated images and real images---on MS-CoCo validation data~\cite{mscoco}.
Table~\ref{table:coco_reward} shows that FID score is significantly reduced when the model is fine-tuned using only human data, despite better image-text alignment.
However, by adding the unlabeled and pre-training datasets, FID score is improved without impacting image-text alignment.
We provide image samples from unseen text prompts in Figure~\ref{fig:abl_fig}. We see that fine-tuned models indeed generate more natural images when exploiting more diverse datasets.


%


\begin{table}[t]
\begin{center}
 \small
\begin{tabular}{ccc}
\toprule
& \multirow{2}{*}{\begin{tabular}[c]{@{}c@{}} FID on \\ {MS-CoCo} ($\downarrow$) \end{tabular}} 
& \multirow{2}{*}{\begin{tabular}[c]{@{}c@{}} Average rewards on \\ tested prompts ($\uparrow$) \end{tabular}}  \\
& & \\ \midrule
Original model & $13.97$  & $0.43$   \\ \midrule
\multirow{2}{*}{\begin{tabular}[c]{@{}c@{}} Fine-tuned model w.o \\   unlabeled \& pre-train \end{tabular}} 
 &\multirow{2}{*}{\begin{tabular}[c]{@{}c@{}}  $26.59$  \end{tabular}}  &\multirow{2}{*}{\begin{tabular}[c]{@{}c@{}}   $0.69$ \end{tabular}}  \\ 
 &&\\ \midrule
 \multirow{2}{*}{\begin{tabular}[c]{@{}c@{}} Fine-tuned model \\  w.o pre-train \end{tabular}} 
 &\multirow{2}{*}{\begin{tabular}[c]{@{}c@{}}  $21.02$  \end{tabular}}  &\multirow{2}{*}{\begin{tabular}[c]{@{}c@{}}   $0.79$ \end{tabular}}  \\
 &&\\  \midrule
Fine-tuned model & $16.76$ &$0.79$   \\
\bottomrule
\end{tabular}
\end{center}
\caption{Comparison with the original Stable Diffusion. For evaluating image fidelity, we measure FID scores on the MS-CoCo. For evaluating the image-text alignment, we measure reward scores and CLIP scores on 120 tested text prompts. $\uparrow(\downarrow)$  indicates that the higher (lower) number is the better.}
\label{table:coco_reward}
\end{table}


%% file: sections/conclusion.tex
\section{Discussion} \label{sec:concl}

In this work, we have demonstrated that fine-tuning with human feedback can effectively improve the image-text alignment in three domains: generating objects with a specified count, color,
or backgrounds.
We analyze several design choices (such as using an auxiliary loss and collecting diverse training data) and find that it is challenging to balance the alignment-fidelity tradeoffs without careful investigations on such design choices.
Even though our results do not address all the failure modes of the existing text-to-image models, 
we hope that our method can serve as a starting point to study learning from human feedback for improving text-to-image models.



{\bf Limitations and future directions}. There are several limitations and interesting future directions in our work:
\begin{itemize} [leftmargin=8mm]
\setlength\itemsep{0.1em}

\item{\em More nuanced human feedback}. Some of the poor generations we observed, such as highly saturated image colors, are likely due to similar images being highly ranked in our training set. We believe that instructing raters to look for a more diverse set of failure modes (oversaturated colors, unrealistic animal anatomy, physics violations, etc.) will improve performance along these axes.

\item {\em Diverse and large human dataset}. For simplicity, we consider a limited class of text categories (count, color, background) and thus consider a simple form of human feedback (good or bad). Due to this, the diversity of our human data is bit limited. 
Extension to more subjective text categories (like artistic generation) and informative human feedback such as ranking would be an important direction for future research.

%

\item {\em Different objectives and algorithms}. For updating the text-to-image model, we use a reward-weighted likelihood maximization. However, similar to prior work in language domains~\cite{instructGPT}, it would be an interesting direction to use RL algorithms~\cite{ppo}. We believe RLHF fine-tuning may lead to better models because (a) it uses online sample generation during updates and (b) KL-regularization over the prior model can mitigate overfitting to the reward function.   
\end{itemize}

%% file: sections/appendix.tex

\section{Qualitative Comparison} \label{app:example}

\begin{figure*} [h!] \centering
\subfigure[Text prompt: {\tt A red colored tiger.}]
{\includegraphics[width=0.99\textwidth]{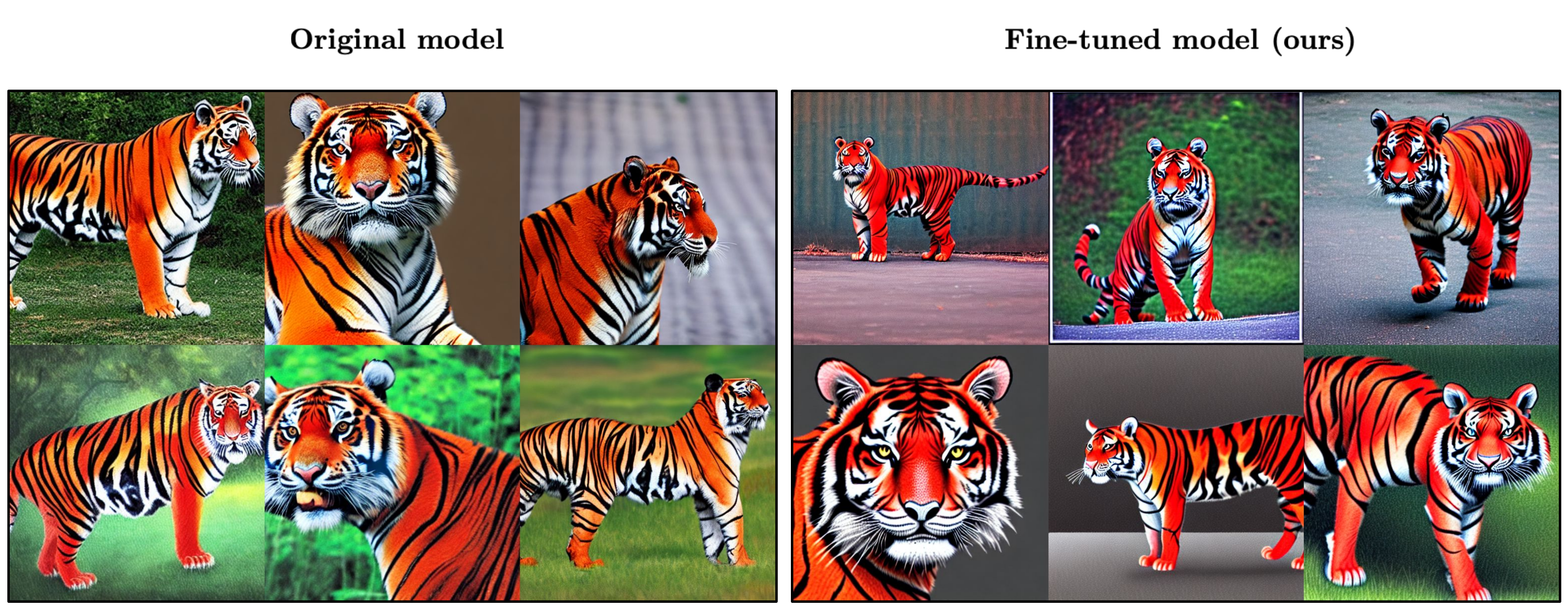}
\label{fig:app_color_0}} 
\subfigure[Text prompt: {\tt A green colored tiger.}]
{\includegraphics[width=0.99\textwidth]{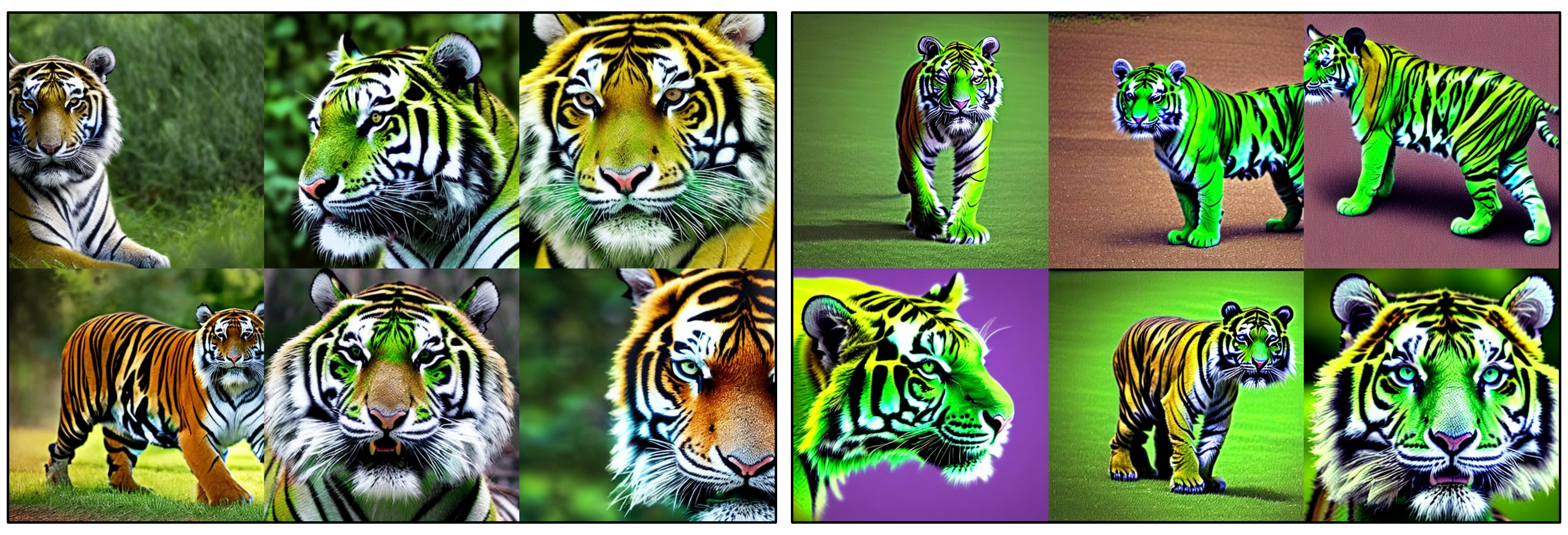}
\label{fig:app_color_1}}
\subfigure[Text prompt: {\tt A pink colored tiger.}]
{\includegraphics[width=0.99\textwidth]{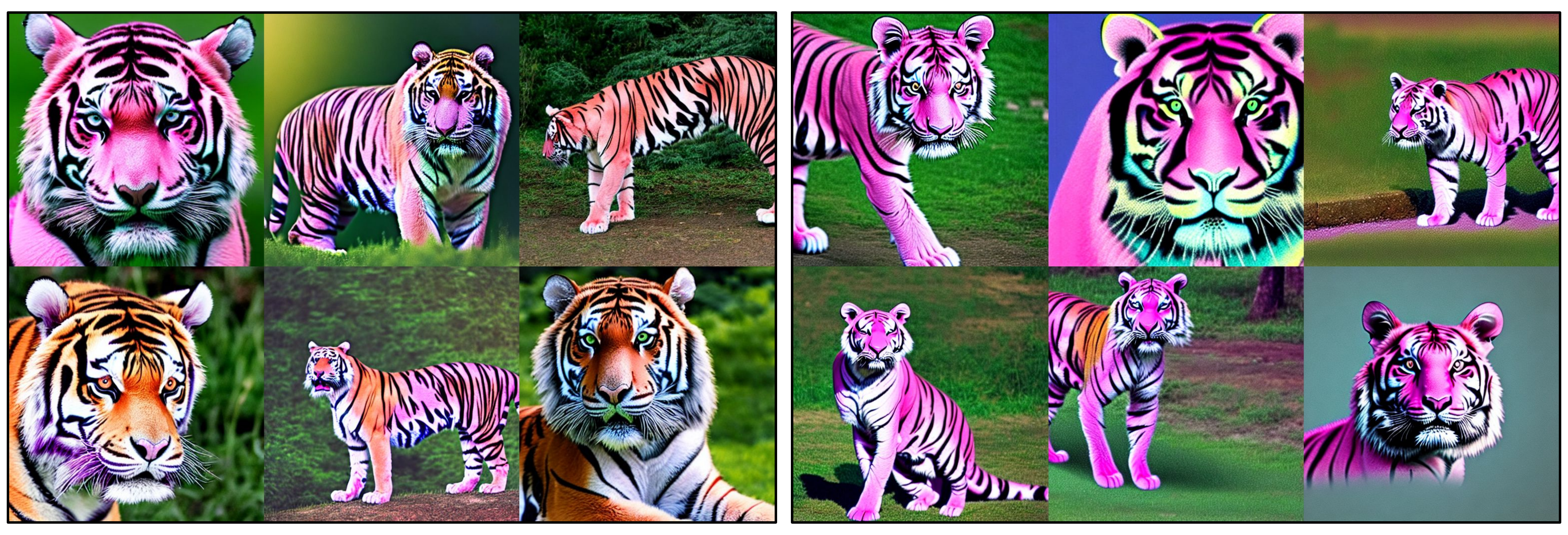}
\label{fig:app_color_2}}
\caption{Samples from the original Stable Diffusion model (left) and our fine-tuned model (right). The fine-tuned model can generate an unseen object ({\tt tiger}) with specified colors.}
\label{fig:app_color}
\end{figure*}

\begin{figure*} [t!] \centering
\subfigure[Text prompt: {\tt Three wolves in the forest.}]
{
\includegraphics[width=0.99\textwidth]{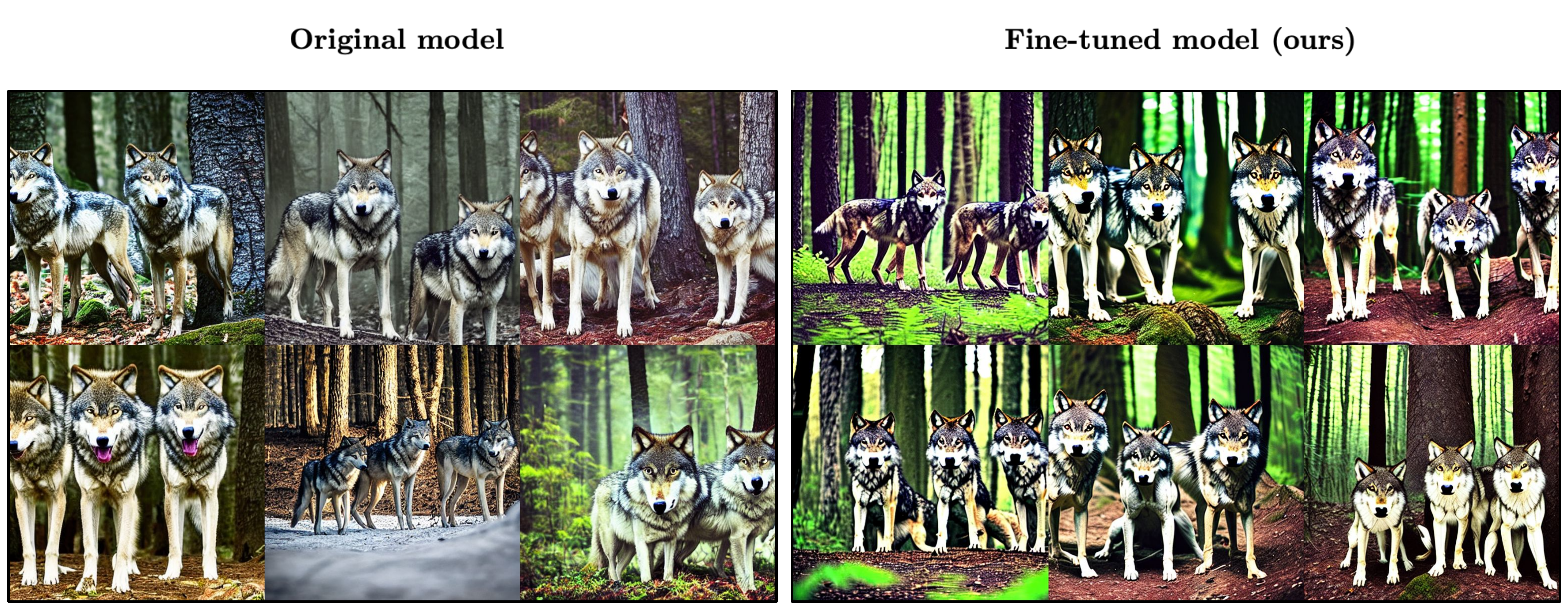}
\label{fig:app_count_0}} 
\subfigure[Text prompt: {\tt Four wolves in the forest.}]
{\includegraphics[width=0.99\textwidth]{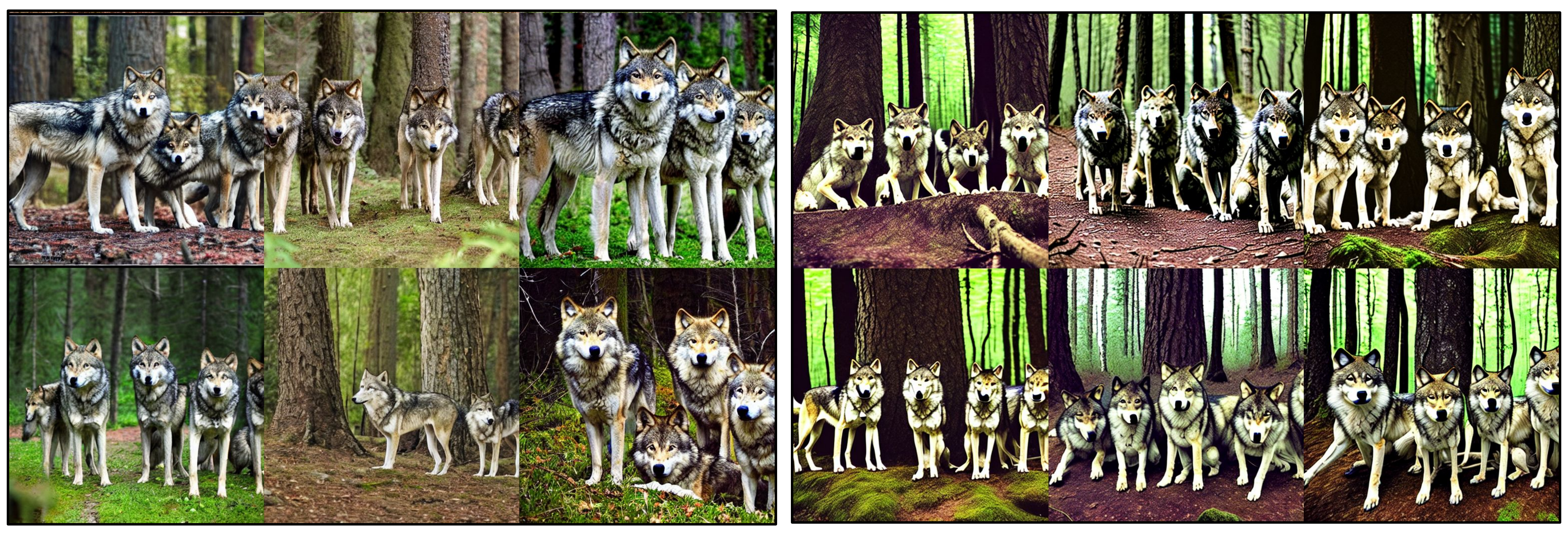}
\label{fig:app_count_1}}
\subfigure[Text prompt: {\tt Five wolves in the forest.}]
{\includegraphics[width=0.99\textwidth]{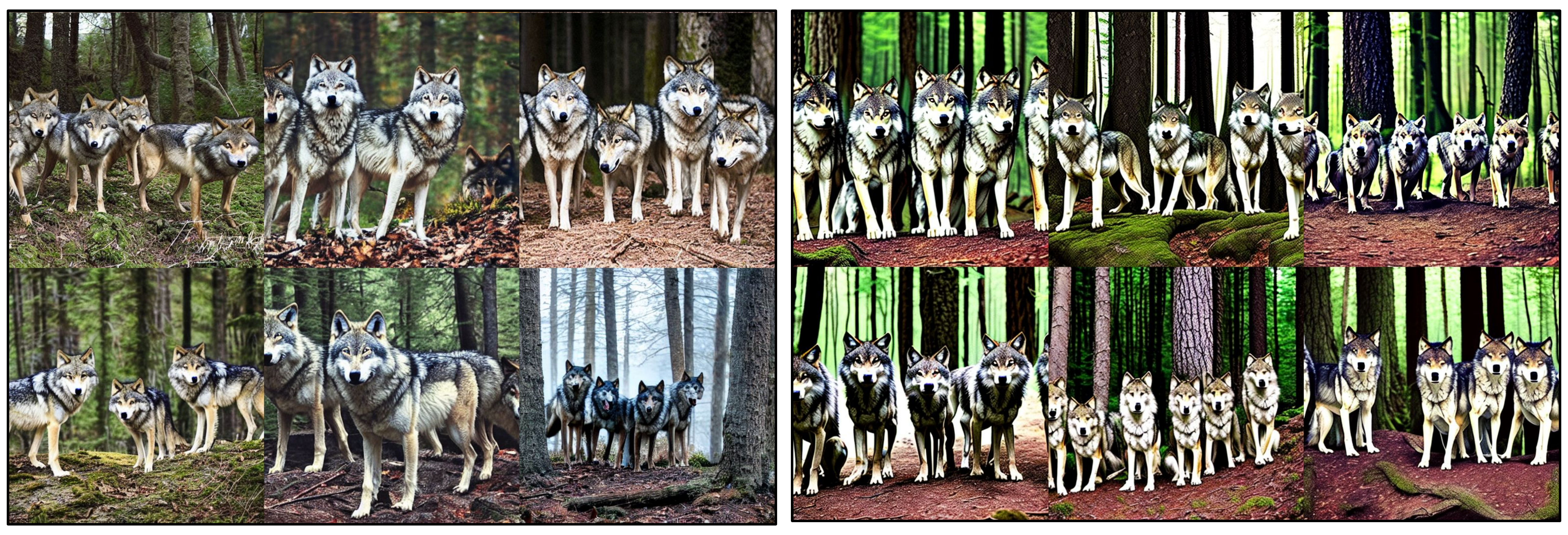}
\label{fig:app_count_2}}
\caption{Samples from the original Stable Diffusion model (left) and our fine-tuned model (right). The fine-tuned model can generate an unseen object ({\tt wolf}) with specified counts. However, the counts are not always perfect, showing a room for improvement.}
\label{fig:app_count}
\end{figure*}

\begin{figure*} [t!] \centering
\subfigure[Text prompt: {\tt A cake in the city.}]
{
\includegraphics[width=0.99\textwidth]{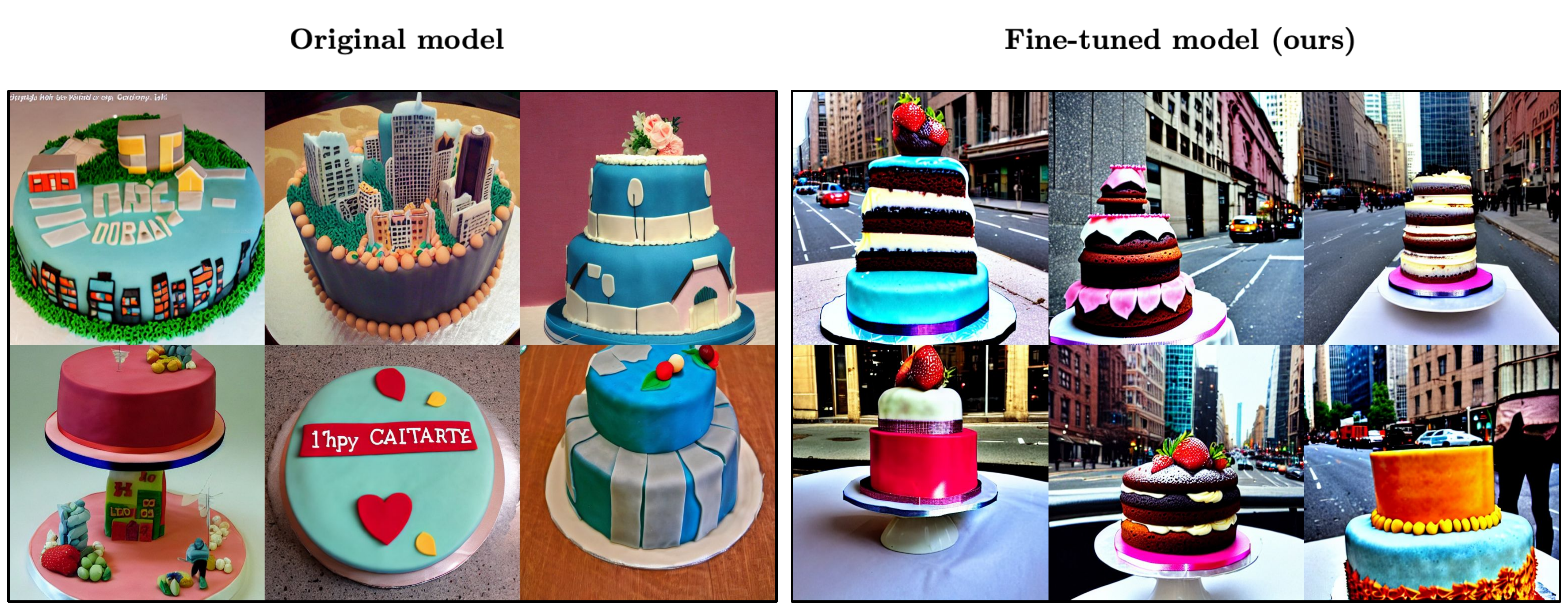}
\label{fig:app_back_0}} 
\subfigure[Text prompt: {\tt A cake in the sea.}]
{\includegraphics[width=0.99\textwidth]{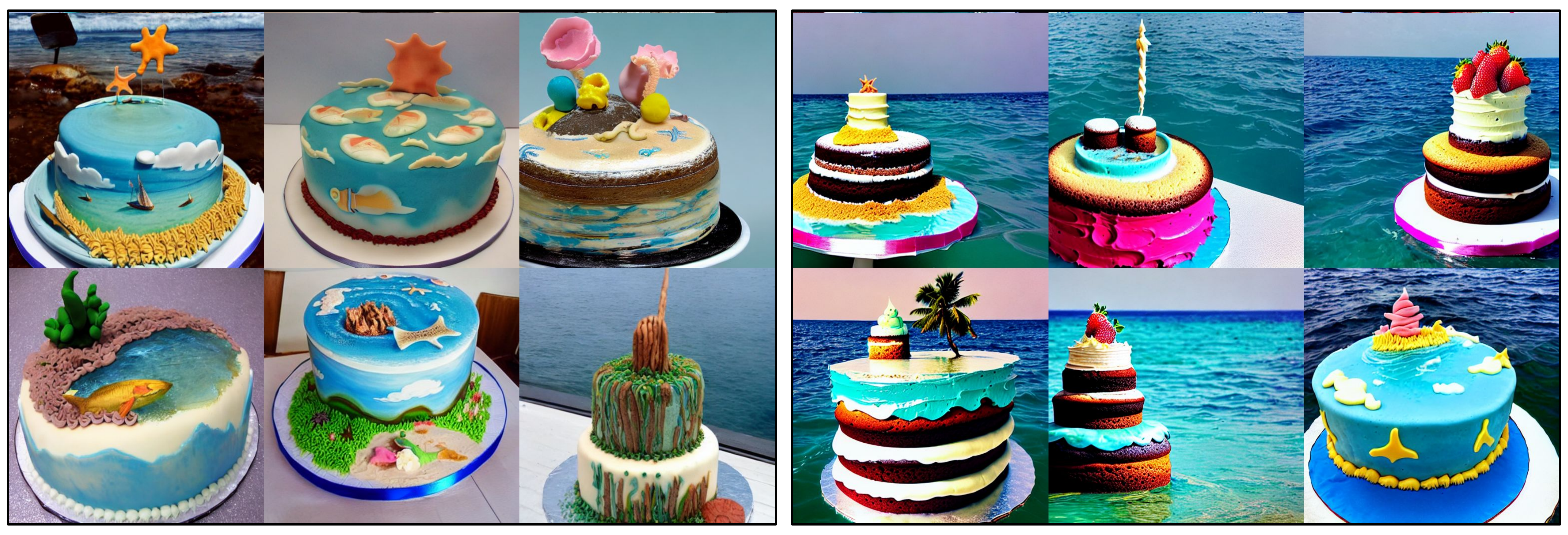}
\label{fig:app_back_1}}
\subfigure[Text prompt: {\tt A cake on the moon.}]
{\includegraphics[width=0.99\textwidth]{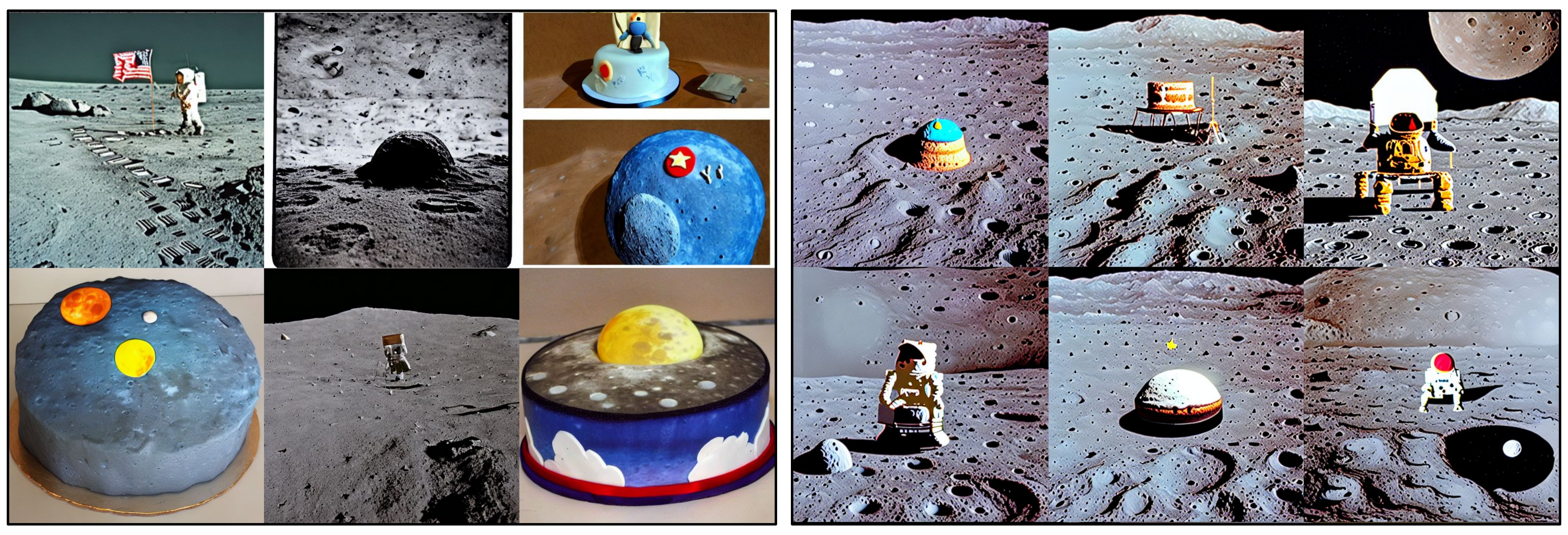}
\label{fig:app_back_2}}
\caption{Samples from the original Stable Diffusion model (left) and our fine-tuned model (right). The fine-tuned model can generate {\tt cake} with specified backgrounds.}
\label{fig:app_back}
\end{figure*}

\begin{figure*} [t!] \centering
\subfigure[Text prompt: {\tt An oil painting of rabbit.}]
{
\includegraphics[width=0.99\textwidth]{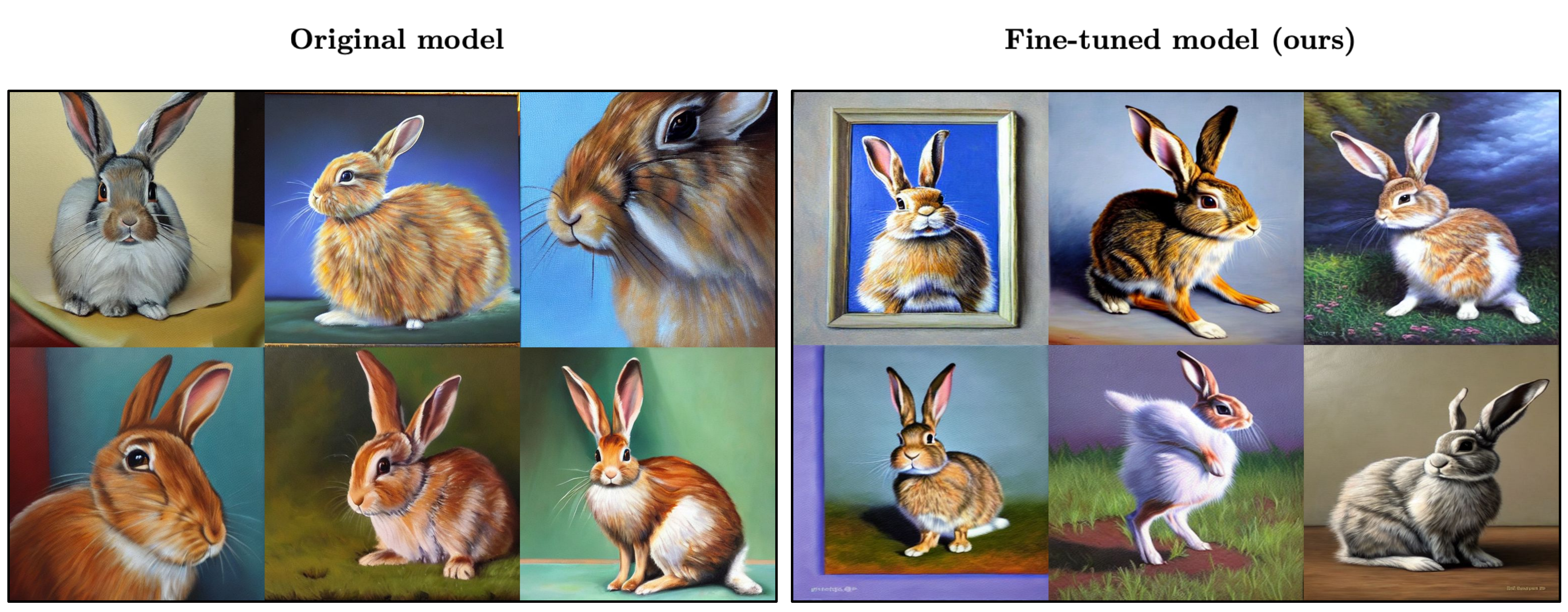}
\label{fig:app_style_0}} 
\subfigure[Text prompt: {\tt A black and white sketch of rabbit.}]
{\includegraphics[width=0.99\textwidth]{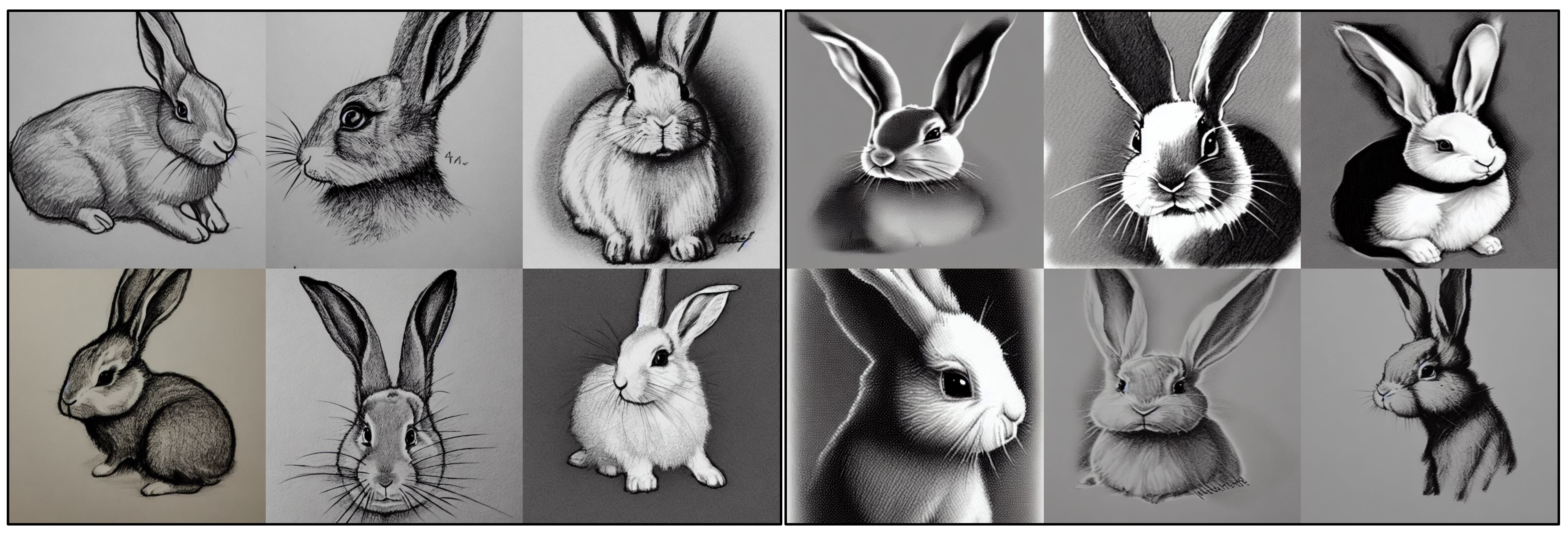}
\label{fig:app_style_1}}
\subfigure[Text prompt: {\tt A 3D render of rabbit.}]
{\includegraphics[width=0.99\textwidth]{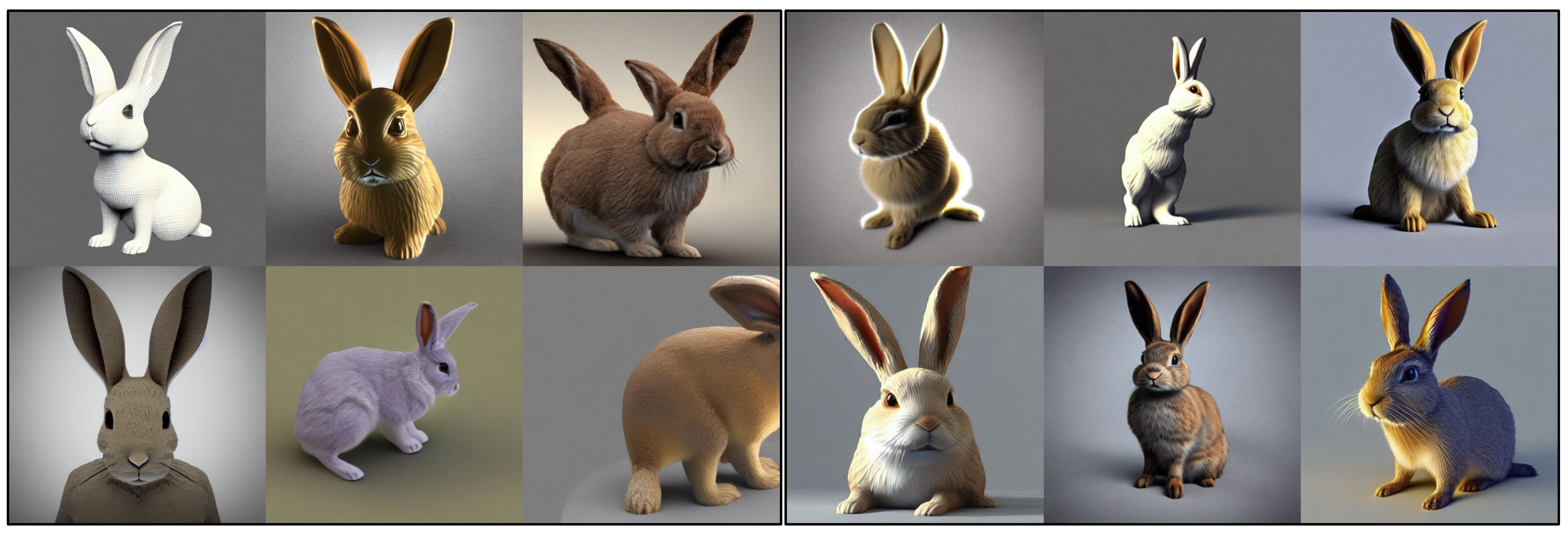}
\label{fig:app_style_2}}
\caption{Samples from the original Stable Diffusion model (left) and our fine-tuned model (right). The fine-tuned model still generate {\tt rabbit} in specified styles.}
\label{fig:app_style}
\end{figure*}

\begin{figure*} [t!] \centering
\subfigure[Text prompt: {\tt A zombie in the style of Picasso.}]
{
\includegraphics[width=0.99\textwidth]{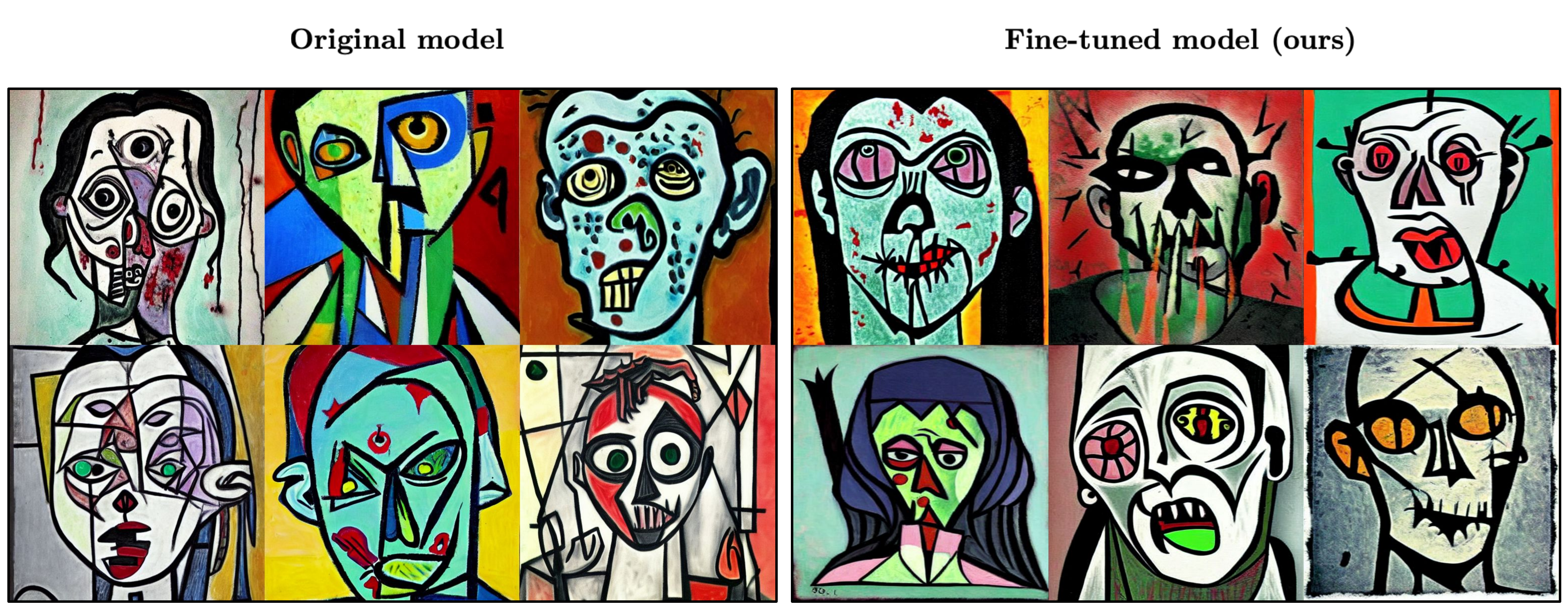}
\label{fig:app_style_new_0}} 
\subfigure[Text prompt: {\tt A watercolor painting of a chair that looks like an octopus.}]
{\includegraphics[width=0.99\textwidth]{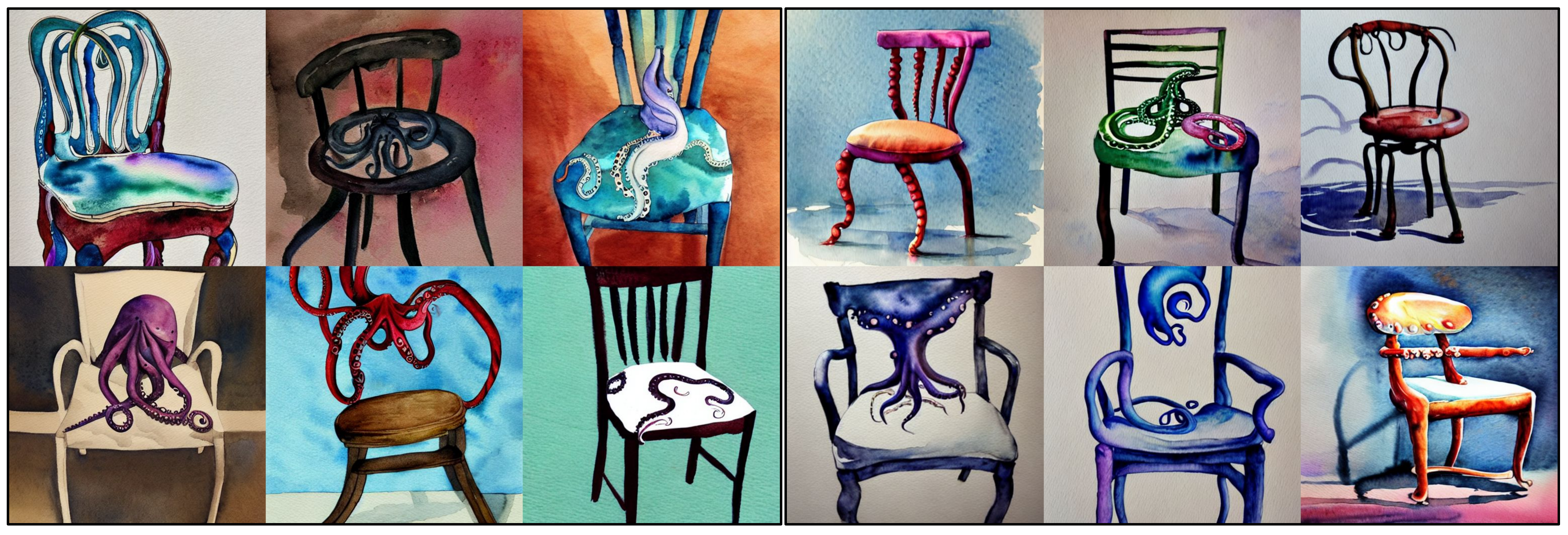}
\label{fig:app_style_new_1}}
\subfigure[Text prompt: {\tt A painting of a squirrel eating a burger.}]
{\includegraphics[width=0.99\textwidth]{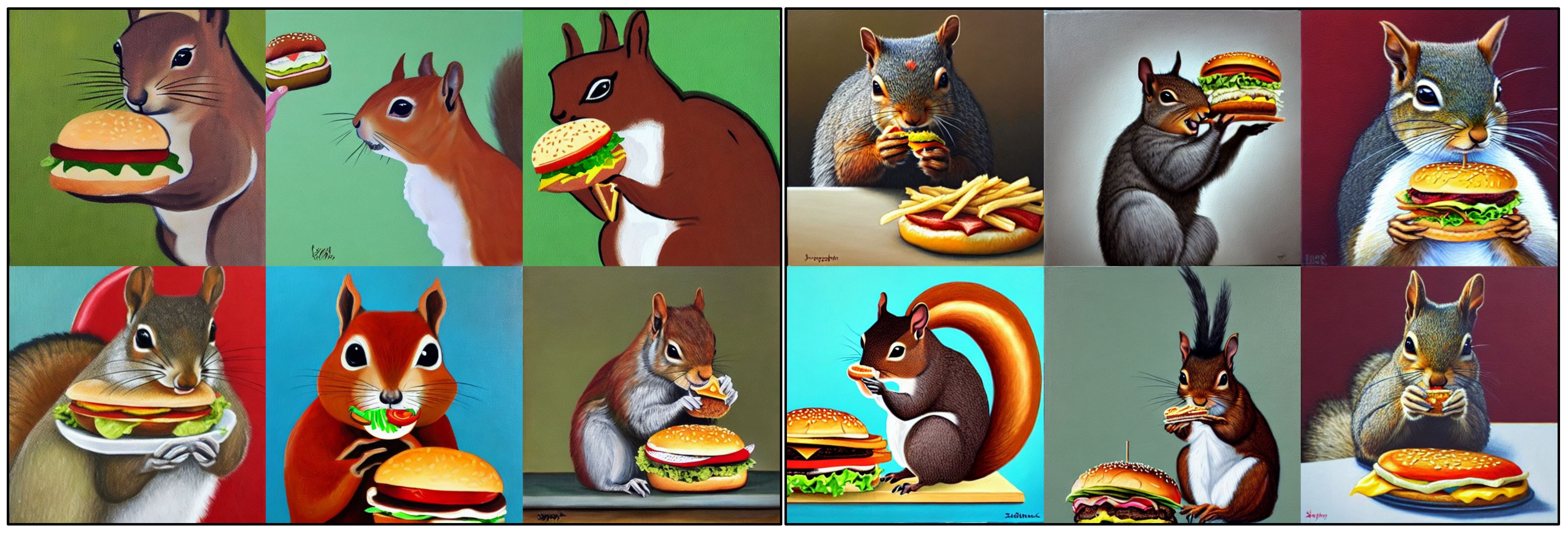}
\label{fig:app_style_new_2}}
\caption{Samples from the original Stable Diffusion model (left) and our fine-tuned model (right). The fine-tuned model still maintains performance across a wide distribution of text prompts.}
\label{fig:app_style_new}
\end{figure*}

\newpage

\section{Image-text Dataset} \label{app:dataset}

In this section, we describe our image-text dataset. We generate 2774 text prompts by combining a word or phrase from that category with some object. Specifically, we consider 9 colors ({\tt red, yellow, green, blue, black, pink, purple, white, brown}), 6 numbers ({\tt 1-6}), 8 backgrounds ({\tt forest, city, moon, field, sea, table, desert, San Franciso}) and 25 objects ({\tt dog, cat, lion, orange, vase, cup, apple, chair, bird, cake, bicycle, tree, donut, box, plate, clock, backpack, car, airplane, bear, horse, tiger, rabbit, rose, wolf}).\footnote{We use the following 5 objects only for evaluation: {\tt bear, tiger, rabbit, rose, wolf}.} 
For each text prompt, we generate 60 or 6 images according to the text category. In total, our image-text dataset consists of 27528 image-text pairs. Labeling for training is done by two human labelers. 

For evaluation, we use 120 text prompts listed in Table~\ref{app:full_list}. 
Given two (anonymized) sets of 4 images, we ask human raters to assess which is better w.r.t.\ image-text alignment and fidelity (i.e., image quality).
Each query is rated by 9 independent human raters in Figure~\ref{fig:main} and Figure~\ref{fig:rejection}.

\begin{table}[h]
\begin{center}
 \small
\begin{tabular}{lc}
\toprule
Category & Examples \\ \midrule
\multirow{2}{*}{\begin{tabular}[c]{@{}c@{}} {Seen} \end{tabular}} 
& A red colored dog.; A red colored donut.; A red colored cake.; A red colored vase.; A green colored dog.;  \\ 
& A green colored donut.; A green colored cake.; A green colored vase.; A pink colored dog.; A pink colored donut.;  \\ 
& A pink colored cake.; A pink colored vase.; A blue colored dog.; A blue colored donut.; A blue colored cake.; \\ 
& A blue colored vase.; A black colored apple.; A green colored apple.; A pink colored apple.; A blue colored apple.; \\
& A dog on the moon.; A donut on the moon.; A cake on the moon.; A vase on the moon.; An apple on the moon.;  \\
& A dog in the sea.; A donut in the sea.; A cake in the sea.; A vase in the sea.; An apple in the sea.; \\
& A dog in the city.; A donut in the city.; A cake in the city.; A vase in the city.; An apple in the city.;  \\
& A dog in the forest.; A donut in the forest.; A cake in the forest.; A vase in the forest.; An apple in the forest.;  \\
& Two dogs.; Two donuts.; Two cakes.; Two vases.; Two apples.; Three dogs.; Three donuts.; Three cakes.; \\
& Three vases.; Three apples.; Four dogs.; Four donuts.; Four cakes.; Four vases.; Four apples.; Five dogs.; \\
\midrule
\multirow{2}{*}{\begin{tabular}[c]{@{}c@{}} {Unseen} \end{tabular}} 
&  A red colored bear.; A red colored wolf.; A red colored tiger.; A red colored rabbit.; A green colored bear.; 
\\
& A green colored wolf.; A green colored tiger.; A green colored rabbit.; A pink colored bear.; A pink colored wolf.;  \\ 
& A pink colored tiger.; A pink colored rabbit.; A blue colored bear.; A blue colored wolf.; A blue colored tiger.; \\
& A blue colored rabbit.; A black colored rose.; A green colored rose.; A pink colored rose.; A blue colored rose.;  \\
& A bear on the moon.; A wolf on the moon.; A tiger on the moon.; A rabbit on the moon.; A rose on the moon.;  \\
& A bear in the sea.; A wolf in the sea.; A tiger in the sea.; A rabbit in the sea.; A rose in the sea.;  \\
& A bear in the city.; A wolf in the city.; A tiger in the city.; A rabbit in the city.; A rose in the city.; \\
& A bear in the forest.; A wolf in the forest.; A tiger in the forest.; A rabbit in the forest.; A rose in the forest.;  \\
& Two brown bears.; Two wolves.; Two tigers.; Two rabbits.; Two red roses.; Three brown bears.; Three wolves.; \\
& Three tigers.; Three rabbits.; Three red roses.; Four brown bears.; Four wolves.; Four tigers.; Four rabbits.; \\
\bottomrule
\end{tabular}
\end{center}
\caption{Examples of text prompts for evaluation.}
\label{app:full_list}
\end{table}

\section{Additional Results}

\begin{table}[h]
\begin{center}
 \small
\begin{tabular}{llll|lll}
\toprule
\multirow{3}{*}{\begin{tabular}[c]{@{}c@{}} {Seen prompts} \end{tabular}} 
& \multicolumn{3}{c}{Image-text alignment} & \multicolumn{3}{c}{Fidelity} \\ \cmidrule(lr){2-7}
&   {Win} & {Lose} & {Tie} &   {Win} & {Lose} & {Tie} \\
\midrule
Color & 58.9 \stdv{19.5}& 10.0 \stdv{7.1}& 31.1 \stdv{26.2}& 53.9 \stdv{39.4}& 28.3 \stdv{24.0}& 17.8 \stdv{23.1} \\
Count  & 69.4 \stdv{6.8}& 11.7 \stdv{5.3}& 18.9 \stdv{10.7}& 42.2 \stdv{36.4}& 37.2 \stdv{23.6}& 20.6 \stdv{22.9}
  \\
Background & 53.3 \stdv{13.3}& 14.4 \stdv{6.4}& 32.2 \stdv{18.0}& 43.9 \stdv{31.0}& 39.4 \stdv{17.1}& 16.7 \stdv{20.1}
  \\ \bottomrule
\multirow{3}{*}{\begin{tabular}[c]{@{}c@{}} {Unseen prompts} \end{tabular}} 
& \multicolumn{3}{c}{Image-text alignment} & \multicolumn{3}{c}{Fidelity} \\ \cmidrule(lr){2-7}
&   {Win} & {Lose} & {Tie} &   {Win} & {Lose} & {Tie} \\
\midrule
Color & 57.2 \stdv{8.5}& 14.4 \stdv{9.6}& 28.3 \stdv{16.3}& 39.4 \stdv{24.7}& 38.9 \stdv{9.7}& 21.7 \stdv{25.1} \\
Count  & 69.4 \stdv{16.1}& 8.3 \stdv{2.4}& 22.2 \stdv{18.0}& 42.8 \stdv{31.8}& 36.7 \stdv{9.1}& 20.6 \stdv{23.3}  \\
Background & 55.6 \stdv{9.6}& 11.1 \stdv{9.9}& 33.3 \stdv{18.9}& 46.1 \stdv{28.5}& 35.6 \stdv{13.6}& 18.3 \stdv{20.7}  \\
\bottomrule
\end{tabular}
\end{center}
\caption{Percentage of generated images from our fine-tuned model that are better than (win), tied with, or worse than (lose) the compared to original stable diffusion model in terms of image-text alignment and fidelity.}
\end{table}

\begin{table}[h]
\begin{center}
 \small
\begin{tabular}{llll|lll}
\toprule
\multirow{3}{*}{\begin{tabular}[c]{@{}c@{}} {Seen prompts} \end{tabular}}
& \multicolumn{3}{c}{Image-text alignment} & \multicolumn{3}{c}{Fidelity} \\ \cmidrule(lr){2-7}
&   {Win} & {Lose} & {Tie} &   {Win} & {Lose} & {Tie} \\
\midrule
Color & 51.7 \stdv{18.6}& 20.6 \stdv{13.2}& 27.8 \stdv{31.1}& 37.8 \stdv{16.0}& 30.6 \stdv{15.2}& 31.7 \stdv{29.7}  \\
Count  & 67.8 \stdv{14.2}& 12.8 \stdv{7.1}& 19.4 \stdv{20.6}& 31.7 \stdv{18.4}& 31.7 \stdv{21.2}& 36.7 \stdv{38.1}
  \\
Background & 47.8 \stdv{15.3}& 10.0 \stdv{6.7}& 42.2 \stdv{20.3}& 41.1 \stdv{21.3}& 27.8 \stdv{14.9}& 31.1 \stdv{34.8}
  \\ \bottomrule
\multirow{3}{*}{\begin{tabular}[c]{@{}c@{}} {Unseen prompts} \end{tabular}} 
& \multicolumn{3}{c}{Image-text alignment} & \multicolumn{3}{c}{Fidelity} \\ \cmidrule(lr){2-7}
&   {Win} & {Lose} & {Tie} &   {Win} & {Lose} & {Tie} \\
\midrule
Color & 47.8 \stdv{7.5}& 20.0 \stdv{14.3}& 32.2 \stdv{20.8}& 31.7 \stdv{21.9}& 39.4 \stdv{13.2}& 28.9 \stdv{32.1} \\
Count & 62.2 \stdv{23.8}& 8.3 \stdv{6.2}& 29.4 \stdv{29.4}& 40.0 \stdv{27.9}& 25.6 \stdv{12.6}& 34.4 \stdv{38.4}  \\
Background & 62.8 \stdv{11.3}& 7.2 \stdv{5.8}& 30.0 \stdv{16.8}& 51.1 \stdv{17.6}& 23.9 \stdv{12.0}& 25.0 \stdv{27.8} \\
\bottomrule
\end{tabular}
\end{center}
\caption{Percentage of generated images from original stable diffusion model with rejection sampling that are better than (win), tied with, or worse than (lose) the compared to original stable diffusion model in terms of image-text alignment and fidelity.}
\end{table}

\begin{table}[h]
\begin{center}
 \small
\begin{tabular}{llll|lll}
\toprule
\multirow{3}{*}{\begin{tabular}[c]{@{}c@{}} {Seen prompts} \end{tabular}}
& \multicolumn{3}{c}{Image-text alignment} & \multicolumn{3}{c}{Fidelity} \\ \cmidrule(lr){2-7}
&   {Win} & {Lose} & {Tie} &   {Win} & {Lose} & {Tie} \\
\midrule
Color & 34.4 \stdv{19.2}& 27.2 \stdv{18.9}& 38.3 \stdv{37.6}& 41.1 \stdv{33.7}& 27.8 \stdv{8.2}& 31.1 \stdv{33.1} \\
Count  & 32.8 \stdv{11.1}& 35.0 \stdv{15.8}& 32.2 \stdv{25.3}& 32.8 \stdv{28.4}& 41.7 \stdv{7.8}& 25.6 \stdv{29.9}
  \\
Background & 31.7 \stdv{12.2}& 24.4 \stdv{16.1}& 43.9 \stdv{27.0}& 36.1 \stdv{24.0}& 36.1 \stdv{15.2}& 27.8 \stdv{35.5}
  \\ \bottomrule
\multirow{3}{*}{\begin{tabular}[c]{@{}c@{}} {Unseen prompts} \end{tabular}} 
& \multicolumn{3}{c}{Image-text alignment} & \multicolumn{3}{c}{Fidelity} \\ \cmidrule(lr){2-7}
&   {Win} & {Lose} & {Tie} &   {Win} & {Lose} & {Tie} \\
\midrule
Color & 40.0 \stdv{19.7}& 23.3 \stdv{12.7}& 36.7 \stdv{32.0}& 39.4 \stdv{20.3}& 32.8 \stdv{14.6}& 27.8 \stdv{32.7}\\
Count  & 53.3 \stdv{15.5}& 27.2 \stdv{9.2}& 19.4 \stdv{23.5}& 26.1 \stdv{19.5}& 52.2 \stdv{7.5}& 21.7 \stdv{25.4} \\
Background & 33.9 \stdv{23.1}& 21.7 \stdv{9.7}& 44.4 \stdv{32.2}& 33.3 \stdv{25.6}& 37.8 \stdv{10.6}& 28.9 \stdv{33.7} \\
\bottomrule
\end{tabular}
\end{center}
\caption{Percentage of generated images from our fine-tuned model that are better than (win), tied with, or worse than (lose) the compared to original stable diffusion model with rejection sampling in terms of image-text alignment and fidelity.}
\end{table}

\section{Experimental Details} \label{app:exp_detail}

{\bf Model architecture}. For our baseline generative model, we use stable diffusion v1.5~\cite{stablediffusion}, which has been pre-trained on large image-text datasets~\cite{laion400m,laion-5b}.
For the reward model, we use ViT-L/14 CLIP model~\cite{clip} to extract image and text embeddings and train a MLP using these embeddings as input. Specifically, we use two-layer MLPs with 1024 hidden dimensions each.
We use ReLUs for the activation function between layers, and we use the Sigmoid activation
function for the output. For auxiliary task, we use temperature $T=2$ and penalty parameter $\lambda=0.5$.

{\bf Training}. Our fine-tuning pipeline is based on publicly released repository (\url{https://github.com/huggingface/diffusers/tree/main/examples/text_to_image}). 
We update the model using AdamW~\cite{adamw} with $\beta_1=0.9$, $\beta_2=0.999$, $\epsilon=1e-8$ and weight decay $1e-2$. The model is trained in half-precision on 4 40GB NVIDIA A100 GPUs, with a per-GPU batch size of 8, resulting in a toal batch size of 512 (256 for pre-training data and 256 for model-generated data).\footnote{In prior work~\cite{instructGPT}, model is optimized with bigger batch for pre-training data. However, in our experiments, using a bigger batch does not make a big difference. We expect this is because small pre-training dataset is used in our work.} It is trained for a total of 10,000 updates.


{\bf FID measurement using MS-CoCo dataset}. We measure FID scores to evaluate the fidelity of different models using MS-CoCo validation dataset (i.e., {\tt val2014}). There are a few caption annotations for each MS-CoCo image. We randomly choose one caption for each image, which results in 40,504 caption and image pairs. MS-CoCo images have different resolutions and they are resized to $256\times256$ before computing FID scores. We use {\tt pytorch-fid} Python implementation for the FID measurement (\url{https://github.com/mseitzer/pytorch-fid}).

\newpage

\section{Pseudocode} \label{app:pseudo}

\begin{algorithm}
\caption{Reward Learning Pseudocode}
\label{alg:reward}
\definecolor{codeblue}{rgb}{0.28125,0.46875,0.8125}
\lstset{
    basicstyle=\fontsize{9pt}{9pt}\ttfamily\bfseries,
    commentstyle=\fontsize{9pt}{9pt}\color{codeblue},
    keywordstyle=
}
\begin{lstlisting}[language=python]
# x, z, y: image, text prompt, human label
# clip: pre-trained CLIP model
# preprocess: Image transform
# pred_r: two-layers MLPs
# Get_perturbated_prompts: function to generate perturbated text prompts
# lambda: penalty parameter
# T: temperature
# N: # of perturbated text prompts

# main model
def RewardFunction(x, z):
    # compute embeddings for tokens
    img_embedding = clip.encode_image(prepocess(x))
    txt_embedding = clip.encode_text(clip.tokenize(z))
    input_embeds = concatenate(img_embedding, txt_embedding)
    
    # predict score
    return pred_r(input_embeds)

# training loop
for (x, z, y) in dataloader: # dims: (batch_size, dim)
    # MSE loss
    r_preds = RewardFunction(x, z)
    loss = MSELoss(r_preds, y)

    # Prompt classification
    scores = [r_preds]
    for z_neg in Get_perturbated_prompts(z, N):
        scores.append(RewardFunction(x, z_neg))
    scores = scores / T
    labels = [0] * batch_size # origin text is always class 0
    loss += lambda * CrossEntropyLoss(scores, labels)

    # update reward function
    optimizer.zero_grad(); loss.backward(); optimizer.step()
\end{lstlisting}
\end{algorithm}

\begin{algorithm}
\caption{Perturbated Text Prompts Generation Pseudocode}
\label{alg:rule}
\definecolor{codeblue}{rgb}{0.28125,0.46875,0.8125}
\lstset{
    basicstyle=\fontsize{9pt}{9pt}\ttfamily\bfseries,
    commentstyle=\fontsize{9pt}{9pt}\color{codeblue},
    keywordstyle=
}
\begin{lstlisting}[language=python]
# z: image, text prompt, human label
# N: # of perturbated text prompts

def Get_perturbated_prompts(z, N):
    color_list = [``red'', ``yellow'', ...]
    obj_list = [``dog'', ``cat'', ...]
    count_list = [``One'', ``Two'', ...]
    loc_list = [``in the sea.'', ``in the sky.'', ...]

    output = []
    count = 0
    while (count < N):
        idx = random.randint(0, len(count_list)-1)
        count = count_list[idx]
        idx = random.randint(0, len(color_list)-1)
        color = color_list[idx]
        idx = random.randint(0, len(loc_list)-1)
        loc = loc_list[idx]
        idx = random.randint(0, len(obj_list)-1)
        obj = obj_list[idx]
        
        if count == ``One'':
            text = ``{} {} {} {}.''.format(count, color, obj, loc)
        else:
            text = ``{} {} {}s {}.''.format(count, color, obj, loc)

        if z != text:
            count += 1
            output.append(text)
    return output
        
\end{lstlisting}
\end{algorithm}





%% file: main.bbl
\begin{thebibliography}{49}
\providecommand{\natexlab}[1]{#1}
\providecommand{\url}[1]{\texttt{#1}}
\expandafter\ifx\csname urlstyle\endcsname\relax
  \providecommand{\doi}[1]{doi: #1}\else
  \providecommand{\doi}{doi: \begingroup \urlstyle{rm}\Url}\fi

\bibitem[Askell et~al.(2021)Askell, Bai, Chen, Drain, Ganguli, Henighan, Jones,
  Joseph, Mann, DasSarma, et~al.]{askell2021general}
Askell, A., Bai, Y., Chen, A., Drain, D., Ganguli, D., Henighan, T., Jones, A.,
  Joseph, N., Mann, B., DasSarma, N., et~al.
\newblock A general language assistant as a laboratory for alignment.
\newblock \emph{arXiv preprint arXiv:2112.00861}, 2021.

\bibitem[Bahdanau et~al.(2016)Bahdanau, Brakel, Xu, Goyal, Lowe, Pineau,
  Courville, and Bengio]{bahdanau2016actor}
Bahdanau, D., Brakel, P., Xu, K., Goyal, A., Lowe, R., Pineau, J., Courville,
  A., and Bengio, Y.
\newblock An actor-critic algorithm for sequence prediction.
\newblock \emph{arXiv preprint arXiv:1607.07086}, 2016.

\bibitem[Bai et~al.(2022{\natexlab{a}})Bai, Jones, Ndousse, Askell, Chen,
  DasSarma, Drain, Fort, Ganguli, Henighan, et~al.]{bai2022training}
Bai, Y., Jones, A., Ndousse, K., Askell, A., Chen, A., DasSarma, N., Drain, D.,
  Fort, S., Ganguli, D., Henighan, T., et~al.
\newblock Training a helpful and harmless assistant with reinforcement learning
  from human feedback.
\newblock \emph{arXiv preprint arXiv:2204.05862}, 2022{\natexlab{a}}.

\bibitem[Bai et~al.(2022{\natexlab{b}})Bai, Kadavath, Kundu, Askell, Kernion,
  Jones, Chen, Goldie, Mirhoseini, McKinnon, et~al.]{bai2022constitutional}
Bai, Y., Kadavath, S., Kundu, S., Askell, A., Kernion, J., Jones, A., Chen, A.,
  Goldie, A., Mirhoseini, A., McKinnon, C., et~al.
\newblock Constitutional ai: Harmlessness from ai feedback.
\newblock \emph{arXiv preprint arXiv:2212.08073}, 2022{\natexlab{b}}.

\bibitem[Brown et~al.(2020)Brown, Mann, Ryder, Subbiah, Kaplan, Dhariwal,
  Neelakantan, Shyam, Sastry, Askell, et~al.]{gpt3}
Brown, T.~B., Mann, B., Ryder, N., Subbiah, M., Kaplan, J., Dhariwal, P.,
  Neelakantan, A., Shyam, P., Sastry, G., Askell, A., et~al.
\newblock Language models are few-shot learners.
\newblock \emph{arXiv preprint arXiv:2005.14165}, 2020.

\bibitem[Christiano et~al.(2017)Christiano, Leike, Brown, Martic, Legg, and
  Amodei]{preference_drl}
Christiano, P.~F., Leike, J., Brown, T., Martic, M., Legg, S., and Amodei, D.
\newblock Deep reinforcement learning from human preferences.
\newblock In \emph{Advances in Neural Information Processing Systems}, 2017.

\bibitem[Cubuk et~al.(2019)Cubuk, Zoph, Mane, Vasudevan, and
  Le]{cubuk2019autoaugment}
Cubuk, E.~D., Zoph, B., Mane, D., Vasudevan, V., and Le, Q.~V.
\newblock Autoaugment: Learning augmentation strategies from data.
\newblock In \emph{Proceedings of the IEEE/CVF Conference on Computer Vision
  and Pattern Recognition}, 2019.

\bibitem[Feng et~al.(2022)Feng, He, Fu, Jampani, Akula, Narayana, Basu, Wang,
  and Wang]{feng2022training}
Feng, W., He, X., Fu, T.-J., Jampani, V., Akula, A., Narayana, P., Basu, S.,
  Wang, X.~E., and Wang, W.~Y.
\newblock Training-free structured diffusion guidance for compositional
  text-to-image synthesis.
\newblock \emph{arXiv preprint arXiv:2212.05032}, 2022.

\bibitem[Gal et~al.(2022)Gal, Alaluf, Atzmon, Patashnik, Bermano, Chechik, and
  Cohen-Or]{gal2022image}
Gal, R., Alaluf, Y., Atzmon, Y., Patashnik, O., Bermano, A.~H., Chechik, G.,
  and Cohen-Or, D.
\newblock An image is worth one word: Personalizing text-to-image generation
  using textual inversion.
\newblock \emph{arXiv preprint arXiv:2208.01618}, 2022.

\bibitem[Goodfellow et~al.(2020)Goodfellow, Pouget-Abadie, Mirza, Xu,
  Warde-Farley, Ozair, Courville, and Bengio]{gan}
Goodfellow, I., Pouget-Abadie, J., Mirza, M., Xu, B., Warde-Farley, D., Ozair,
  S., Courville, A., and Bengio, Y.
\newblock Generative adversarial networks.
\newblock \emph{Communications of the ACM}, 63\penalty0 (11):\penalty0
  139--144, 2020.

\bibitem[Hessel et~al.(2021)Hessel, Holtzman, Forbes, Bras, and
  Choi]{hessel2021clipscore}
Hessel, J., Holtzman, A., Forbes, M., Bras, R.~L., and Choi, Y.
\newblock Clipscore: A reference-free evaluation metric for image captioning.
\newblock \emph{arXiv preprint arXiv:2104.08718}, 2021.

\bibitem[Heusel et~al.(2017)Heusel, Ramsauer, Unterthiner, Nessler, and
  Hochreiter]{fid}
Heusel, M., Ramsauer, H., Unterthiner, T., Nessler, B., and Hochreiter, S.
\newblock Gans trained by a two time-scale update rule converge to a local nash
  equilibrium.
\newblock In \emph{Advances in neural information processing systems}, 2017.

\bibitem[Ho et~al.(2020)Ho, Jain, and Abbeel]{ho2020denoising}
Ho, J., Jain, A., and Abbeel, P.
\newblock Denoising diffusion probabilistic models.
\newblock In \emph{Advances in Neural Information Processing Systems}, 2020.

\bibitem[Ibarz et~al.(2018)Ibarz, Leike, Pohlen, Irving, Legg, and
  Amodei]{ibarz2018preference_demo}
Ibarz, B., Leike, J., Pohlen, T., Irving, G., Legg, S., and Amodei, D.
\newblock Reward learning from human preferences and demonstrations in atari.
\newblock In \emph{Advances in Neural Information Processing Systems}, 2018.

\bibitem[Kingma \& Welling(2013)Kingma and Welling]{vae}
Kingma, D.~P. and Welling, M.
\newblock Auto-encoding variational bayes.
\newblock \emph{arXiv preprint arXiv:1312.6114}, 2013.

\bibitem[Kreutzer et~al.(2018)Kreutzer, Khadivi, Matusov, and
  Riezler]{kreutzer2018can}
Kreutzer, J., Khadivi, S., Matusov, E., and Riezler, S.
\newblock Can neural machine translation be improved with user feedback?
\newblock \emph{arXiv preprint arXiv:1804.05958}, 2018.

\bibitem[Krizhevsky et~al.(2017)Krizhevsky, Sutskever, and
  Hinton]{krizhevsky2017imagenet}
Krizhevsky, A., Sutskever, I., and Hinton, G.~E.
\newblock Imagenet classification with deep convolutional neural networks.
\newblock \emph{Communications of the ACM}, 60\penalty0 (6):\penalty0 84--90,
  2017.

\bibitem[Kumari et~al.(2022)Kumari, Zhang, Zhang, Shechtman, and
  Zhu]{kumari2022multi}
Kumari, N., Zhang, B., Zhang, R., Shechtman, E., and Zhu, J.-Y.
\newblock Multi-concept customization of text-to-image diffusion.
\newblock \emph{arXiv preprint arXiv:2212.04488}, 2022.

\bibitem[Lee et~al.(2021)Lee, Smith, and Abbeel]{pebble}
Lee, K., Smith, L., and Abbeel, P.
\newblock Pebble: Feedback-efficient interactive reinforcement learning via
  relabeling experience and unsupervised pre-training.
\newblock In \emph{International Conference on Machine Learning}, 2021.

\bibitem[Lin et~al.(2014)Lin, Maire, Belongie, Hays, Perona, Ramanan,
  Doll{\'a}r, and Zitnick]{mscoco}
Lin, T.-Y., Maire, M., Belongie, S., Hays, J., Perona, P., Ramanan, D.,
  Doll{\'a}r, P., and Zitnick, C.~L.
\newblock Microsoft coco: Common objects in context.
\newblock In \emph{European conference on computer vision}, 2014.

\bibitem[Liu et~al.(2023)Liu, Sferrazza, and Abbeel]{liu2023chain}
Liu, H., Sferrazza, C., and Abbeel, P.
\newblock Chain of hindsight aligns language models with feedback.
\newblock \emph{arXiv preprint arXiv: Arxiv-2302.02676}, 2023.

\bibitem[Liu et~al.(2022{\natexlab{a}})Liu, Li, Du, Torralba, and
  Tenenbaum]{liu2022compositional}
Liu, N., Li, S., Du, Y., Torralba, A., and Tenenbaum, J.~B.
\newblock Compositional visual generation with composable diffusion models.
\newblock \emph{arXiv preprint arXiv:2206.01714}, 2022{\natexlab{a}}.

\bibitem[Liu et~al.(2022{\natexlab{b}})Liu, Garrette, Saharia, Chan, Roberts,
  Narang, Blok, Mical, Norouzi, and Constant]{liu2022character}
Liu, R., Garrette, D., Saharia, C., Chan, W., Roberts, A., Narang, S., Blok,
  I., Mical, R., Norouzi, M., and Constant, N.
\newblock Character-aware models improve visual text rendering.
\newblock \emph{arXiv preprint arXiv:2212.10562}, 2022{\natexlab{b}}.

\bibitem[Loshchilov \& Hutter(2017)Loshchilov and Hutter]{adamw}
Loshchilov, I. and Hutter, F.
\newblock Decoupled weight decay regularization.
\newblock \emph{arXiv preprint arXiv:1711.05101}, 2017.

\bibitem[MacGlashan et~al.(2017)MacGlashan, Ho, Loftin, Peng, Roberts, Taylor,
  and Littman]{coach}
MacGlashan, J., Ho, M.~K., Loftin, R., Peng, B., Roberts, D., Taylor, M.~E.,
  and Littman, M.~L.
\newblock Interactive learning from policy-dependent human feedback.
\newblock In \emph{International Conference on Machine Learning}, 2017.

\bibitem[Madhyastha et~al.(2019)Madhyastha, Wang, and
  Specia]{madhyastha2019vifidel}
Madhyastha, P., Wang, J., and Specia, L.
\newblock Vifidel: Evaluating the visual fidelity of image descriptions.
\newblock \emph{arXiv preprint arXiv:1907.09340}, 2019.

\bibitem[Nakano et~al.(2021)Nakano, Hilton, Balaji, Wu, Ouyang, Kim, Hesse,
  Jain, Kosaraju, Saunders, et~al.]{webgpt}
Nakano, R., Hilton, J., Balaji, S., Wu, J., Ouyang, L., Kim, C., Hesse, C.,
  Jain, S., Kosaraju, V., Saunders, W., et~al.
\newblock Webgpt: Browser-assisted question-answering with human feedback.
\newblock \emph{arXiv preprint arXiv:2112.09332}, 2021.

\bibitem[Ouyang et~al.(2022)Ouyang, Wu, Jiang, Almeida, Wainwright, Mishkin,
  Zhang, Agarwal, Slama, Ray, et~al.]{instructGPT}
Ouyang, L., Wu, J., Jiang, X., Almeida, D., Wainwright, C.~L., Mishkin, P.,
  Zhang, C., Agarwal, S., Slama, K., Ray, A., et~al.
\newblock Training language models to follow instructions with human feedback.
\newblock \emph{arXiv preprint arXiv:2203.02155}, 2022.

\bibitem[Radford et~al.(2021)Radford, Kim, Hallacy, Ramesh, Goh, Agarwal,
  Sastry, Askell, Mishkin, Clark, et~al.]{clip}
Radford, A., Kim, J.~W., Hallacy, C., Ramesh, A., Goh, G., Agarwal, S., Sastry,
  G., Askell, A., Mishkin, P., Clark, J., et~al.
\newblock Learning transferable visual models from natural language
  supervision.
\newblock In \emph{International Conference on Machine Learning}, 2021.

\bibitem[Raffel et~al.(2020)Raffel, Shazeer, Roberts, Lee, Narang, Matena,
  Zhou, Li, Liu, et~al.]{t5}
Raffel, C., Shazeer, N., Roberts, A., Lee, K., Narang, S., Matena, M., Zhou,
  Y., Li, W., Liu, P.~J., et~al.
\newblock Exploring the limits of transfer learning with a unified text-to-text
  transformer.
\newblock \emph{J. Mach. Learn. Res.}, 21\penalty0 (140):\penalty0 1--67, 2020.

\bibitem[Ramesh et~al.(2021)Ramesh, Pavlov, Goh, Gray, Voss, Radford, Chen, and
  Sutskever]{dalle1}
Ramesh, A., Pavlov, M., Goh, G., Gray, S., Voss, C., Radford, A., Chen, M., and
  Sutskever, I.
\newblock Zero-shot text-to-image generation.
\newblock In \emph{International Conference on Machine Learning}, 2021.

\bibitem[Ramesh et~al.(2022)Ramesh, Dhariwal, Nichol, Chu, and Chen]{dalle2}
Ramesh, A., Dhariwal, P., Nichol, A., Chu, C., and Chen, M.
\newblock Hierarchical text-conditional image generation with clip latents.
\newblock \emph{arXiv preprint arXiv:2204.06125}, 2022.

\bibitem[Rombach et~al.(2022)Rombach, Blattmann, Lorenz, Esser, and
  Ommer]{stablediffusion}
Rombach, R., Blattmann, A., Lorenz, D., Esser, P., and Ommer, B.
\newblock High-resolution image synthesis with latent diffusion models.
\newblock In \emph{Proceedings of the IEEE/CVF Conference on Computer Vision
  and Pattern Recognition}, 2022.

\bibitem[Ruiz et~al.(2022)Ruiz, Li, Jampani, Pritch, Rubinstein, and
  Aberman]{dreambooth}
Ruiz, N., Li, Y., Jampani, V., Pritch, Y., Rubinstein, M., and Aberman, K.
\newblock Dreambooth: Fine tuning text-to-image diffusion models for
  subject-driven generation.
\newblock \emph{arXiv preprint arXiv:2208.12242}, 2022.

\bibitem[Saharia et~al.(2022)Saharia, Chan, Saxena, Li, Whang, Denton,
  Ghasemipour, Ayan, Mahdavi, Lopes, et~al.]{imagen}
Saharia, C., Chan, W., Saxena, S., Li, L., Whang, J., Denton, E., Ghasemipour,
  S. K.~S., Ayan, B.~K., Mahdavi, S.~S., Lopes, R.~G., et~al.
\newblock Photorealistic text-to-image diffusion models with deep language
  understanding.
\newblock In \emph{Advances in Neural Information Processing Systems}, 2022.

\bibitem[Scheurer et~al.(2022)Scheurer, Campos, Chan, Chen, Cho, and
  Perez]{scheurer2022training}
Scheurer, J., Campos, J.~A., Chan, J.~S., Chen, A., Cho, K., and Perez, E.
\newblock Training language models with language feedback.
\newblock \emph{arXiv preprint arXiv: Arxiv-2204.14146}, 2022.

\bibitem[Schuhmann et~al.(2021)Schuhmann, Vencu, Beaumont, Kaczmarczyk, Mullis,
  Katta, Coombes, Jitsev, and Komatsuzaki]{laion400m}
Schuhmann, C., Vencu, R., Beaumont, R., Kaczmarczyk, R., Mullis, C., Katta, A.,
  Coombes, T., Jitsev, J., and Komatsuzaki, A.
\newblock Laion-400m: Open dataset of clip-filtered 400 million image-text
  pairs.
\newblock \emph{arXiv preprint arXiv:2111.02114}, 2021.

\bibitem[Schuhmann et~al.(2022)Schuhmann, Beaumont, Vencu, Gordon, Wightman,
  Cherti, Coombes, Katta, Mullis, Wortsman, et~al.]{laion-5b}
Schuhmann, C., Beaumont, R., Vencu, R., Gordon, C., Wightman, R., Cherti, M.,
  Coombes, T., Katta, A., Mullis, C., Wortsman, M., et~al.
\newblock Laion-5b: An open large-scale dataset for training next generation
  image-text models.
\newblock \emph{arXiv preprint arXiv:2210.08402}, 2022.

\bibitem[Schulman et~al.(2017)Schulman, Wolski, Dhariwal, Radford, and
  Klimov]{ppo}
Schulman, J., Wolski, F., Dhariwal, P., Radford, A., and Klimov, O.
\newblock Proximal policy optimization algorithms.
\newblock \emph{arXiv preprint arXiv:1707.06347}, 2017.

\bibitem[Sohl-Dickstein et~al.(2015)Sohl-Dickstein, Weiss, Maheswaranathan, and
  Ganguli]{sohl2015deep}
Sohl-Dickstein, J., Weiss, E., Maheswaranathan, N., and Ganguli, S.
\newblock Deep unsupervised learning using nonequilibrium thermodynamics.
\newblock In \emph{International Conference on Machine Learning}, 2015.

\bibitem[Stiennon et~al.(2020)Stiennon, Ouyang, Wu, Ziegler, Lowe, Voss,
  Radford, Amodei, and Christiano]{stiennon2020learning}
Stiennon, N., Ouyang, L., Wu, J., Ziegler, D.~M., Lowe, R., Voss, C., Radford,
  A., Amodei, D., and Christiano, P.
\newblock Learning to summarize from human feedback.
\newblock \emph{arXiv preprint arXiv:2009.01325}, 2020.

\bibitem[Van Den~Oord et~al.(2016)Van Den~Oord, Kalchbrenner, and
  Kavukcuoglu]{van2016pixel}
Van Den~Oord, A., Kalchbrenner, N., and Kavukcuoglu, K.
\newblock Pixel recurrent neural networks.
\newblock In \emph{International conference on machine learning}, pp.\
  1747--1756, 2016.

\bibitem[Warnell et~al.(2018)Warnell, Waytowich, Lawhern, and Stone]{deeptamer}
Warnell, G., Waytowich, N., Lawhern, V., and Stone, P.
\newblock Deep tamer: Interactive agent shaping in high-dimensional state
  spaces.
\newblock In \emph{Conference on Artificial Intelligence}, 2018.

\bibitem[Wu et~al.(2021)Wu, Ouyang, Ziegler, Stiennon, Lowe, Leike, and
  Christiano]{wu2021recursively}
Wu, J., Ouyang, L., Ziegler, D.~M., Stiennon, N., Lowe, R., Leike, J., and
  Christiano, P.
\newblock Recursively summarizing books with human feedback.
\newblock \emph{arXiv preprint arXiv:2109.10862}, 2021.

\bibitem[Xue et~al.(2022)Xue, Barua, Constant, Al-Rfou, Narang, Kale, Roberts,
  and Raffel]{xue2022byt5}
Xue, L., Barua, A., Constant, N., Al-Rfou, R., Narang, S., Kale, M., Roberts,
  A., and Raffel, C.
\newblock Byt5: Towards a token-free future with pre-trained byte-to-byte
  models.
\newblock \emph{Transactions of the Association for Computational Linguistics},
  10:\penalty0 291--306, 2022.

\bibitem[Yu et~al.(2022{\natexlab{a}})Yu, Wang, Vasudevan, Yeung,
  Seyedhosseini, and Wu]{yu2022coca}
Yu, J., Wang, Z., Vasudevan, V., Yeung, L., Seyedhosseini, M., and Wu, Y.
\newblock Coca: Contrastive captioners are image-text foundation models.
\newblock \emph{arXiv preprint arXiv:2205.01917}, 2022{\natexlab{a}}.

\bibitem[Yu et~al.(2022{\natexlab{b}})Yu, Xu, Koh, Luong, Baid, Wang,
  Vasudevan, Ku, Yang, Ayan, et~al.]{parti}
Yu, J., Xu, Y., Koh, J.~Y., Luong, T., Baid, G., Wang, Z., Vasudevan, V., Ku,
  A., Yang, Y., Ayan, B.~K., et~al.
\newblock Scaling autoregressive models for content-rich text-to-image
  generation.
\newblock \emph{arXiv preprint arXiv:2206.10789}, 2022{\natexlab{b}}.

\bibitem[Zhou \& Xu(2020)Zhou and Xu]{zhou2020learning}
Zhou, W. and Xu, K.
\newblock Learning to compare for better training and evaluation of open domain
  natural language generation models.
\newblock In \emph{Conference on Artificial Intelligence}, 2020.

\bibitem[Ziegler et~al.(2019)Ziegler, Stiennon, Wu, Brown, Radford, Amodei,
  Christiano, and Irving]{ziegler2019fine}
Ziegler, D.~M., Stiennon, N., Wu, J., Brown, T.~B., Radford, A., Amodei, D.,
  Christiano, P., and Irving, G.
\newblock Fine-tuning language models from human preferences.
\newblock \emph{arXiv preprint arXiv:1909.08593}, 2019.

\end{thebibliography}
